\documentclass[letterpaper, 10 pt, conference]{ieeeconf}  
\IEEEoverridecommandlockouts                              
\usepackage{url}
\usepackage{amsmath,amssymb}
\usepackage{graphicx,color}
\usepackage{booktabs}
\usepackage{multirow}
\usepackage{verbatim}
\usepackage{gensymb}
\let\labelindent\relax  
\usepackage{enumitem}
\usepackage{pifont}
\usepackage{xspace}
\usepackage{adjustbox}
\graphicspath{{figures/}}

\newcommand{\ba}{\mathbf{a}}

\newcommand{\bs}{\mathbf{s}}

\newcommand{\ie}{\emph{i.e.,}\xspace}
\newcommand{\eg}{\emph{e.g.,}\xspace}
\newcommand{\alg}{DCUR\xspace}
\newcommand{\tbuf}{\mathcal{D}^{(T)}}
\newcommand{\sbuf}{\mathcal{D}^{(S)}}
\usepackage{array}
\makeatletter
\g@addto@macro{\endtabular}{\rowfont{}}
\makeatother
\newcommand{\rowfonttype}{}
\newcommand{\rowfont}[1]{
   \gdef\rowfonttype{#1}#1%
}
\newcolumntype{L}{>{\rowfonttype}l}
\newcolumntype{R}{>{\rowfonttype}r}

\usepackage[font={footnotesize}]{caption}
\def\tablename{Table}

\title{\LARGE \bf
DCUR: Data Curriculum for Teaching\\
via Samples with Reinforcement Learning
}

\author{
  Daniel Seita, Abhinav Gopal, Zhao Mandi, John Canny\\
  Department of EECS\\
  University of California Berkeley \\
  \texttt{\{seita, abhinavg, mandi.zhao, canny\}@berkeley.edu} \\
}

\author{Daniel Seita$^{1}$, Abhinav Gopal$^{1}$, Zhao Mandi$^1$, John Canny$^1$
\thanks{$^{1}$University of California, Berkeley, USA.}%
\thanks{Correspondence to {\tt\small seita@berkeley.edu}}%
}
\begin{document}

\maketitle
\thispagestyle{empty}
\pagestyle{empty}


\begin{abstract}
Deep reinforcement learning (RL) has shown great empirical successes, but suffers from brittleness and sample inefficiency. A potential remedy is to use a previously-trained policy as a source of supervision. In this work, we refer to these policies as \emph{teachers} and study how to transfer their expertise to new \emph{student} policies by focusing on data usage. We propose a framework, \textbf{D}ata \textbf{CU}rriculum for \textbf{R}einforcement learning (\alg), which first trains teachers using online deep RL, and stores the logged environment interaction history. Then, students learn by running either offline RL or by using teacher data in combination with a small amount of self-generated data. \alg's central idea involves defining a class of data curricula which, as a function of training time, limits the student to sampling from a fixed subset of the full teacher data. We test teachers and students using state-of-the-art deep RL algorithms across a variety of data curricula. Results suggest that the choice of data curricula significantly impacts student learning, and that it is beneficial to limit the data during early training stages while gradually letting the data availability grow over time. We identify when the student can learn offline and match teacher performance without relying on specialized offline RL algorithms. Furthermore, we show that collecting a small fraction of online data provides complementary benefits with the data curriculum. Supplementary material is available at \url{https://tinyurl.com/teach-dcur}. 
\end{abstract}


\section{Introduction}\label{sec:intro}

Humans often learn best when guided through a curricula. When providing expert demonstrations to human students, the demonstrations should fall within a particular range of difficulties. If they are too easy the student learns nothing, but if they are too difficult the student may have trouble learning~\cite{chaiklin,vygotski}. 
With this intuition, we consider the analogous context in Reinforcement Learning (RL)~\cite{Sutton_2018}, and study how a teacher policy can best ``teach'' a student policy, where both are trained with reinforcement learning. We investigate when a teacher provides the student with a dataset of samples (\ie data tuples) $\tbuf$. For the student, rather than propose a new algorithm to learn from data, we use \emph{existing} algorithms but focus on \emph{data usage}. In particular, given a student employing a standard off-the-shelf RL algorithm, how can it better sample from $\tbuf$?

\begin{figure*}[t]
\center
\includegraphics[width=0.92\textwidth]{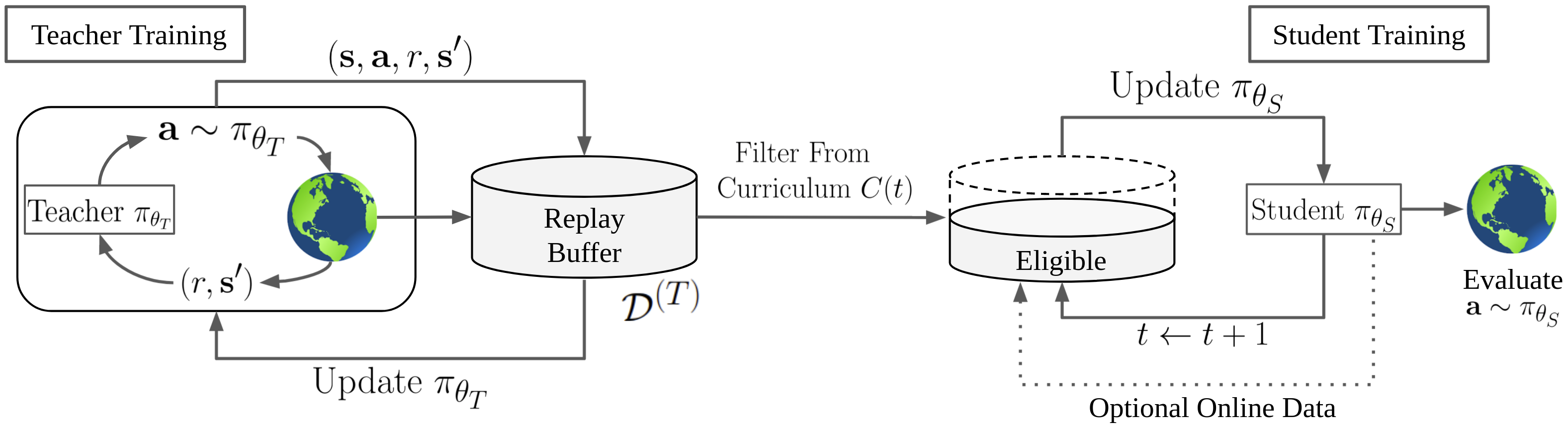}
\caption{
Visualization of \alg. First, a \emph{teacher} policy $\pi_{\theta_T}$ is trained using standard Deep RL with online environment interactions, and fills a replay buffer $\tbuf$ with all the experienced data tuples. After training $\pi_{\theta_T}$, we choose a fixed data curriculum $C(t)$ (see Figure~\ref{fig:method-1}) for training a \emph{student} policy $\pi_{\theta_S}$ from $\tbuf$. In general, we study when students train \emph{offline}, but since this can be challenging, we optionally allow students to gather some online data to use for learning, in addition to $\tbuf$. For each time $t$ (\ie minibatch gradient update), the fixed curriculum strategy $C(t)$ restricts the samples the student can draw from $\tbuf$, resulting in a set of eligible data tuples (shaded in light gray) that the student can use for gradient updates. 
} 
\vspace*{-10pt}
\label{fig:teaser}
\end{figure*}

We propose a framework, \textbf{\alg} (pronounced \emph{dee-curr}): \textbf{D}ata \textbf{CU}rriculum for \textbf{R}einforcement learning, where we study how to filter a teacher's static dataset to accelerate a student's learning progress, measured in terms of environment episodic reward.
The framework is compatible with pure offline reinforcement learning~\cite{batch_rl_ernst_2005,batch_rl_2012,offline_rl_2020} and when the student can engage in a small amount of self-generated data, which we refer to as apprenticeship learning. Either case reduces the need for the student to engage in extensive and potentially risky exploratory behavior, and thus may hold tremendous promise for enabling robots to learn from existing, massive datasets.
In experiments, we test the \alg framework by training teachers in a standard fashion with online environment interaction. We store the logged history of experienced environment interactions as a large dataset $\tbuf$ to be used by the student for learning. 
Results over a range of standard continuous control tasks~\cite{openaigym,mujoco_2012} suggest that students running off-the-shelf, off-policy RL algorithms can make use of curricula to more efficiently learn from static teacher data, even without the use of specialized offline RL algorithms. See Figure~\ref{fig:teaser} for an overview of \alg.
The contributions of this paper include:

\begin{itemize}[leftmargin=*]
    \item We introduce the \alg framework which considers data improvements, rather than algorithmic improvements, for accelerating learning from large, teacher-provided datasets.
    \item We show that the best curricula, combined with potentially longer training, can enable students training offline to match the teacher's top performance.
    \item We show that students can use some online data (2.5-5.0\% of the offline data size) in combination with data curricula to aid learning in more complex environments.
\end{itemize}


\section{Related Work}\label{sec:rw}


\textbf{Curriculum Learning}. The use of curriculum learning in machine learning dates to at least Elman~et~al.~\cite{Elman1993}, who showed the benefit of initially training on ``easier'' samples while gradually increasing the difficulty. Subsequent work by Bengio~et~al.~\cite{bengio_2009} confirmed these results by accelerating classification and language modeling by arranging samples in order of difficulty. Other work in curriculum learning has included training teacher agents to provide samples or loss functions to a student~\cite{l2t_2018,l2t_dlf}, and larger-scale studies to investigate curricula for image classification~\cite{when_curricula_work_2021}. In RL, curriculum learning has shown promise in multi-task~\cite{teacher-student-curriculum-2017,CARML_2019} and self-play~\cite{self-play_2018} contexts, for selecting one of several teachers to provide samples~\cite{seita_zpd_2019}, and for generating a curriculum of start~\cite{reverse_curriculum} or goal~\cite{auto_goal,value_disagreement_2020} states. In this work, we design a curriculum of samples (\ie data tuples) for (mostly) offline RL, and we do not focus on the multi-task setting nor do we require self-play or goal generation. When students execute online steps, this can be interpreted as a form of apprenticeship learning~\cite{apprenticeship_2004} where the student can ``practice'' in addition to using offline data.

\textbf{Offline Reinforcement Learning.} Offline RL~\cite{offline_rl_2020}, also referred to as Batch RL~\cite{batch_rl_ernst_2005,batch_rl_2012}, has seen an explosion of recent interest. Offline RL is the special case of reinforcement learning~\cite{Sutton_2018} without exploration, so the agent must learn from a static dataset. It differs from imitation learning~\cite{algo_imit_2018} in that data is annotated with rewards, which can be utilized by reinforcement learning algorithms to learn better policies than the underlying data generating policy. Many widely utilized Deep RL algorithms, such as DQN~\cite{mnih-dqn-2015,van2016deep} for discrete control and DDPG~\cite{lillicrap2015continuous}, SAC~\cite{sac}, and TD3~\cite{td3} for continuous control are off-policy algorithms and, in principle, capable of learning offline. In practice, however, researchers have found that such off-policy algorithms are highly susceptible to bootstrapping errors and thus may diverge quickly and perform poorly~\cite{BCQ_2019,benchmarking_batchrl_2019,discor_2020,BEAR_2019,implicit_under_2021}. One remedy is to incorporate conservatism such as by regularizing the value functions in model-free RL settings~\cite{BCQ_2019,BEAR_2019,CQL_2020,plas_2020,keep_doing_worked_2020,behavior_regularized_2019,CRR_2020}. Other studies have found promising results in model-based contexts~\cite{MOPO_2020,MOREL_2020,COMBO_2021}. The goals of this work are orthogonal to work that attempts to develop specialized offline RL algorithms, because the focus here is on knowledge transfer from a teacher to a student, with the offline setting as one possible learning scenario for the student.

\textbf{Reinforcement Learning from Teachers.} Combining reinforcement learning with data from teachers is a highly effective technique for training students, particularly for hard exploration environments. One line of research has explored distillation techniques~\cite{actor_mimic_2016,pd_2016,distilling_pd_2019,dual_policy_distillation} for multi-task learning, which trains student networks to match output from teacher networks (\eg Q-values). Another active area of research focuses on demonstrations~\cite{ARGALL2009469}; in off-policy RL, a replay buffer~\cite{Lin1992} can contain teacher demonstrations, which can be used along with self-generated samples from a student. Examples of such algorithms in discrete control settings include DQfD~\cite{dqfd}, Ape-X DQfD~\cite{ape-x_dqfd}, and R2D3~\cite{r2d3}. Other work utilizes demonstrations in continuous control by adding transitions to a replay buffer~\cite{overcoming,leveraging_demos_2017,socket_insertion_2018}, specifying an auxiliary loss~\cite{kickstarting2018} or estimating value functions for model-based RL~\cite{SAVED_2020}. These works enable additional exploration from the student, allowing for self-generated samples, whereas we aim to understand how well students can learn with minimal exploration. Furthermore, these works often use a very low \emph{demo ratio}, or the fraction of expert (teacher) demonstrations in a given minibatch. For example, R2D3~\cite{r2d3} reported that a demo ratio as small as $1/256$ was ideal. That these algorithms perform best when utilizing so little teacher data motivates the need to understand how to use teacher data without requiring frequent student environment interaction.


\begin{figure*}[t]
\center
\includegraphics[width=0.90\textwidth]{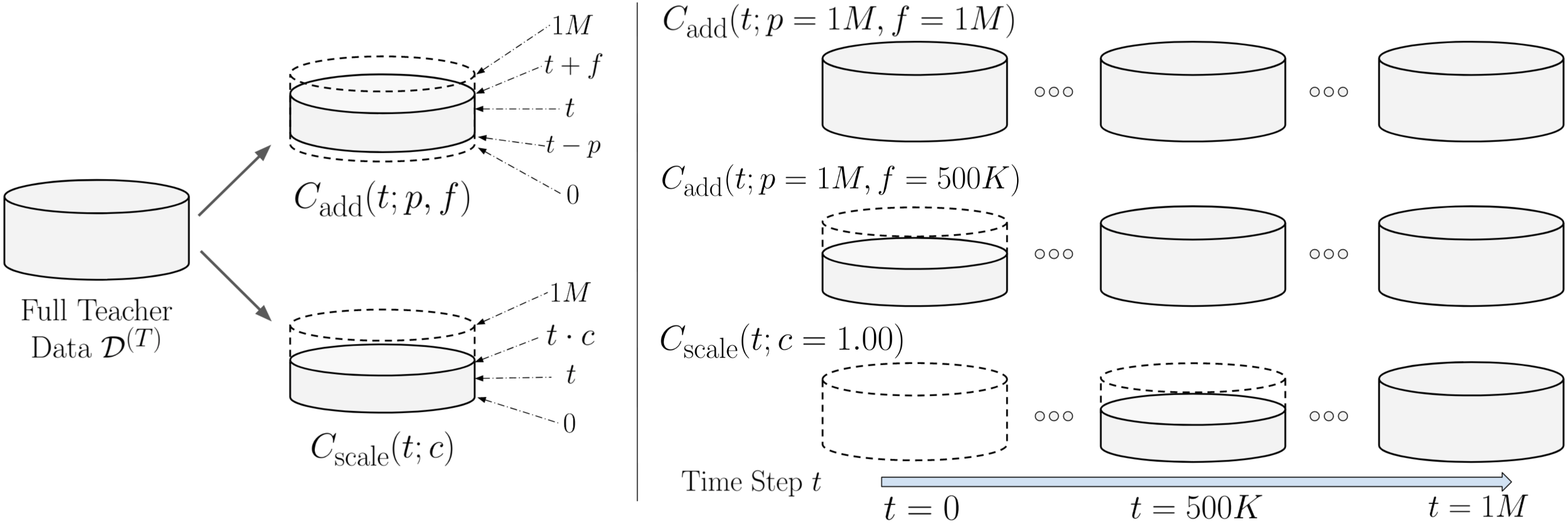}
\caption{
Visualization of the data curricula described in Section~\ref{ssec:curricula}. For all buffer visuals, the bottom represents index 0, \ie the first data tuple from the teacher's training history, and the top is index 1M. At a student training step $t$, data tuples available for sampling are shaded gray.
\textbf{Left}: we illustrate the $C_{\rm add}(t)$ and $C_{\rm scale}(t)$ curricula, which determine different ranges of data tuples in $\tbuf$.
\textbf{Right}: three examples of curricula (one per row) showing how the available data to the student changes over 1M training time steps. For $C_{\rm add}(t; p=1M, f=1M)$, the full $\tbuf$ buffer is available at all times, whereas $C_{\rm scale}(t; c=1.00)$ only enables indices 0 to $t$ for time $t$, and hence the available samples grows linearly throughout training. 
}
\vspace*{-10pt}
\label{fig:method-1}
\end{figure*}

\section{Problem Statement and Preliminaries}\label{sec:ps}

We utilize the Markov Decision Process (MDP) framework for RL~\cite{Sutton_2018}. An MDP is specified as a tuple $(\mathcal{S}, \mathcal{A}, P, R, \gamma)$ where at each time step $t$, the agent is at state $\bs_t \in \mathcal{S}$ and executes action $\ba_t \in \mathcal{A}$. The dynamics map the state-action pair into a successor state ${\bs_{t+1} \sim P(\cdot \;|\; \bs_t, \ba_t)}$, and the agent receives a scalar reward ${r_t = R(\bs_t, \ba_t)}$.
The objective is to find a policy ${\pi : \mathcal{S} \to \mathcal{A}}$ that maximizes the expected discounted return $\mathbb{E}[\sum_{t=0}^\infty \gamma^t r_t]$ with discount $\gamma \in (0,1]$. In Deep RL, the policy $\pi_\theta$ is parameterized by a deep neural network with parameters $\theta$.

The RL framework we study involves two agents: a \emph{teacher} $T$ and a \emph{student} $S$, following respective policies $\pi_{\theta_T}$ and $\pi_{\theta_S}$ with parameters $\theta_T$ and $\theta_S$. We assume the teacher is trained via standard online RL, and produces a dataset $\tbuf = \{ (\bs_i,\ba_i,r_i,\bs_{i+1}) \}_{i=0}^N$ of $N$ \emph{data tuples} (also referred to as ``samples''), where each contains a state $\bs_i$, action $\ba_i$, scalar reward $r_i$, and successor state $\bs_{i+1}$.
In general, we use the $i$ subscript notation in $(\bs_i,\ba_i,r_i,\bs_{i+1})$ to specify a time indexing of the tuples across the full data, and use $(\bs,\ba,r,\bs')$ when precise time indexing is not needed. Data tuples from teacher data $\tbuf$ are provided to the student $S$, which runs an off-policy RL algorithm, so that it can in principle learn from just the fixed data.
We consider the problem of designing a curriculum to decide, for each time step $t$ of the student's learning progress,\footnote{In this work, we consider RL contexts where it is standard to have a 1:1 ratio of environment steps and gradient (\ie training) updates, modulo any initial online data collection to partially fill in a replay buffer before training begins. We thus treat a ``time step'' $t$ as referring to environment steps and gradient updates interchangeably. If students learn offline, then a ``time step'' is interpreted as a gradient update only.} 
which data tuples from $\tbuf$ should be ``available'' to the student when it samples minibatches for gradient updates.


\section{Method}\label{sec:method}

\subsection{Teachers and Data Generation}\label{ssec:teachers}

The \alg framework is agnostic to the precise algorithm to train students and teachers. Unless stated otherwise, we generate teacher data $\tbuf$ using 
TD3~\cite{td3}, a state-of-the-art off-policy Deep RL algorithm for continuous control.
TD3 is an actor-critic algorithm where the actor $\pi_{\theta_T}$ is a policy, and the critic is a value function which consists of two Q-networks, $Q_{\phi_1}$ and $Q_{\phi_2}$, with target networks $Q_{\phi_1, \rm targ}, Q_{\phi_2, \rm targ}$. During gradient updates, TD3 mitigates overestimation of Q-values by taking the minimum of the two Q-networks to compute targets $y$ for the Bellman update:
\begin{equation}\label{eq:target-td3}
y = r + \gamma \min_{i\in \{1,2\}} Q_{\phi_i, \rm targ}(\bs', \ba'(\bs'))
\end{equation}
with discount factor $\gamma$, and where $\ba'$ is the action considered at the successor state $\bs'$:
\begin{equation}
\begin{split}
\ba'(\bs') &= {\rm clip}(\pi_{\theta_T}(\bs') + \epsilon, \ba_{\rm low}, \ba_{\rm high}) \\
& \epsilon  \sim {\rm clip}( \mathcal{N}(0, \sigma), -\beta, \beta )
\end{split}
\end{equation}
which in practice involves adding zero-mean clipped Gaussian noise $\epsilon$ to $\pi_{\theta_T}(\bs')$ for some $\beta$, then clipping (again) component-wise to an environment-dependent action range $[\ba_{\rm low}$, $\ba_{\rm high}]$. For more details on TD3, we refer the reader to Fujimoto~et~al.~\cite{td3}. To generate $\tbuf$, we use the \emph{logged environment interaction history} of the TD3 teacher from online training, resulting in a set of ordered tuples:
\begin{equation}\label{eq:data-logged}
    \tbuf = \{ (\bs_i,\ba_i,r_i,\bs_{i+1}) \}_{i=0}^{N=1M},
\end{equation}
in a \emph{replay buffer}, where following standard MuJoCo training~\cite{td3}, the number of environment steps and the buffer capacity are both 1M, so no data tuples are overwritten. To our knowledge, the only prior work that has tested TD3 in offline RL with logged data is from Agarwal~et~al.~\cite{Optimistic_Perspective_2020}, who report that TD3 outperformed Batch Constrained Q-learning~\cite{BCQ_2019} when learning from logged data generated from DDPG agents. We perform a deeper investigation of training on logged data by showing the utility of curricula (Section~\ref{ssec:curricula}) and self-generated data (Section~\ref{ssec:online-learning}).

\begin{table*}[t]
 \setlength\tabcolsep{5.0pt}
 \caption{
 \textbf{Offline RL Results with \alg}. Student performance as a function of 11 data curricula, with teacher performance (bottom row) as a reference. We report the ``M1'' and ``M2'' evaluation metrics (Section~\ref{sec:experiments}). We run 1 teacher per environment, then use 5 random seeds for training students from that same teacher data. Hence, all numbers reported below for student data curriculum experiments are averages over 5 independent offline RL runs; see the Appendix for standard error values. For each column, values are bolded for the best M1 and M2 results among \emph{all} 11 curricula for students, \emph{and} for other students with overlapping standard errors. The bolded values do not consider teacher performance, which is only present as a reference.
 }
 \centering
 \footnotesize
 \begin{tabular}{@{}lrrrrrrrr@{}} 
 \toprule
 & \multicolumn{2}{c}{Ant-v3} & \multicolumn{2}{c}{HalfCheetah-v3} & \multicolumn{2}{c}{Hopper-v3} & \multicolumn{2}{c}{Walker2d-v3}  \\
 \cmidrule(lr){2-3} \cmidrule(lr){4-5} \cmidrule(lr){6-7} \cmidrule(lr){8-9}
 Data Curriculum & M1 & M2 & M1 & M2 & M1 & M2 & M1 & M2 \\
 \midrule
 ${C_{\rm add}(t ; f=50K)}$  &   219.9 & 382.0 & \textbf{8067.3} & \textbf{7147.0} & 2986.2 & 1669.5 & \textbf{2715.6} & \textbf{1712.1} \\
 ${C_{\rm add}(t ; f=100K)}$ &  -604.2 &  -409.6 & 7416.2 & \textbf{7052.6} & 2627.4 & 1980.1 & \textbf{2746.7} & \textbf{1810.5} \\
 ${C_{\rm add}(t ; f=200K)}$ &   288.4 &  -620.4 & \textbf{8028.4} & 6521.5 & 2525.3 & 1887.9 & \textbf{2691.8} & \textbf{1651.9} \\
 ${C_{\rm add}(t ; f=500K)}$ & 283.3 &  -801.1 & 7108.7 & 5041.2 & 2498.0 & 1546.2 & 1515.7 & 1183.5 \\
 ${C_{\rm add}(t ; f=1M)}$   & -1552.1 & -1390.5 & 7447.3 & 3846.8 & 2323.0 & 1610.3 & 1741.4 & 1096.4 \\
 ${C_{\rm add}(t ; p=800K,f=0)}$  & -889.7 & 1497.9 & \textbf{7650.4} & 6942.7 & 2132.2 & 1924.4 &  792.0 & 1521.8 \\
 \midrule
 ${C_{\rm scale}(t ; c=0.50)}$    &   285.0 &  301.4  & 7467.8 & 6261.9 & 2332.3 & 1450.5 & 1866.0 &  799.9 \\
 ${C_{\rm scale}(t ; c=0.75)}$    &   167.0 &  412.2  & 7392.6 & 6689.2 & \textbf{3284.7} & 1818.2 & 1864.9 & 1251.5 \\
 ${C_{\rm scale}(t ; c=1.00)}$    &   825.6 & 1103.7  & \textbf{8305.3} & 6980.0 & 2984.3 & 2092.6 & 2178.5 & 1305.7 \\
 ${C_{\rm scale}(t ; c=1.10)}$    & \textbf{2952.5} & \textbf{2212.0} & \textbf{8306.0} & \textbf{7095.6} & \textbf{3185.2} & \textbf{2317.1} & 2423.9 & \textbf{1698.1} \\
 ${C_{\rm scale}(t ; c=1.25)}$    & -1851.1 &  199.6  & 7843.8 & \textbf{7175.4} & 2755.5 & 1891.9 & \textbf{2839.4} & \textbf{1747.4} \\
 \midrule
 TD3 Teacher     & 4876.2 & 3975.8 & 8573.6 & 7285.5 & 3635.2 & 2791.9 & 3927.9 & 2579.8 \\
 \bottomrule \vspace{-0.5em} \\
 \end{tabular}
 \label{tab:results-250}
 \vspace{-2.0em}
\end{table*}

\subsection{Data CUrriculum for Reinforcement Learning (\alg)}\label{ssec:curricula}

We propose to accelerate student learning with a curriculum $C(t)$ which specifies the range of eligible data tuples in the static teacher data $\tbuf$ which can be sampled for the minibatch gradient update at time $t$.
We define two classes of curricula: \emph{additive} and \emph{scale}. An additive curricula $C_{\rm add}$ uses two parameters, $p\ge 0$ and $f\ge 0$, specifying the \emph{previous} and \emph{future} data tuples in $\tbuf$ \emph{relative to $t$}.
For ease of notation, we omit the $p$ parameter if the intent is to always allow data tuples from index 0, which represents the teacher's earliest environment interaction. We thus denote the additive curricula on $\tbuf$ using one of two conventions:
\begin{equation}\label{eq:c_add}
\begin{split}
C_{\rm add}(t ; p,f) &= \{ (\bs_i,\ba_i,r_i,\bs_{i+1}) \}_{i=\max(0, t-p)}^{i=\min(1M, t+f)} \\
C_{\rm add}(t ; f) &= \{ (\bs_i,\ba_i,r_i,\bs_{i+1}) \}_{i=0}^{i=\min(1M, t+f)}
\end{split}
\end{equation}
where the valid time indices are centered at the current student training time cursor $t$, and are always limited by the buffer data size of $|\tbuf| = 1M$ studied in this work. The second class of curricula $C_{\rm scale}$, is parameterized by a single \emph{scale} parameter $c>0$ and defined as:
\begin{equation}\label{eq:c_scale}
C_{\rm scale}(t ; c) = \{ (\bs_i,\ba_i,r_i,\bs_{i+1}) \}_{i=0}^{i=\min(1M, t \cdot c)}
\end{equation}
which enables index 0 up to index $t\cdot c$.
See Figure~\ref{fig:method-1} for a visualization. 
If the student trains for 1M gradient updates following ${C_{\rm add}(t ; f=1M)}$, it has the full buffer accessible for sampling data tuples at all times.
Using ${C_{\rm scale}(t ; c=1.00)}$ means the student can only sample from indices 0 up to $t$ at time $t$, so the available offline data grows over time. In addition, ${C_{\rm add}(t ; f=0)}$ and ${C_{\rm scale}(t ; c=1.00)}$ define the same curriculum; we default to the latter notation.

In prior work, Fujimoto~et~al.~\cite{BCQ_2019} tested two special cases of these curricula, named \emph{final buffer} and \emph{concurrent}. The final buffer setting enables the entire data at all times for learning and was later studied in Agarwal~et~al.~\cite{Optimistic_Perspective_2020}. This is equivalent to ${C_{\rm add}(t ; f=1M)}$. 
The concurrent setting is represented as $C_{\rm add}(t; c=1.00)$.
We remark that that limiting the size of the buffer is a known option to stabilize online RL~\cite{deeper_look}, but we aim to study this in an largely offline context in more complex environments, and where the eligible data buffer can grow over time.

\subsection{Apprenticeship Learning: Small Amount of Online Data}\label{ssec:online-learning}

While we primarily study the student learning purely offline, we additionally explore learning with small amounts of on-policy student data, to see how much this stabilizes training and to also check that such effects are complementary with a data curriculum. This means the student forms a smaller buffer $\sbuf$ of self-generated online data with exploration noise, and where $|\sbuf| \ll |\tbuf|$.
We call this setting \emph{X\% Online} if the student, by the end of its training, has collected self-generated data tuples that amount to X\% of the full teacher data (1M in this work). The student still performs 1M gradient updates, but takes one online step at equally spaced time intervals (\ie one step every 100/X updates). For example, with ``5\% Online,'' the student takes online environment steps 1 out of every 20 gradient updates, and $\sbuf$ contains 50K data tuples at the end of training.
This can equivalently be viewed as the student performing 50K consecutive online environment steps, but with 20 gradient updates between consecutive steps.
The data tuples from $\sbuf$ are never discarded. When the student samples a minibatch at time $t$, it first applies the pre-selected curriculum (see Section~\ref{ssec:curricula}) to get the appropriate subset of $\tbuf$, then combines the resulting eligible data tuples with all of $\sbuf$, then uniformly samples from that.


\begin{figure*}[t]
\center
\includegraphics[width=0.85\textwidth]{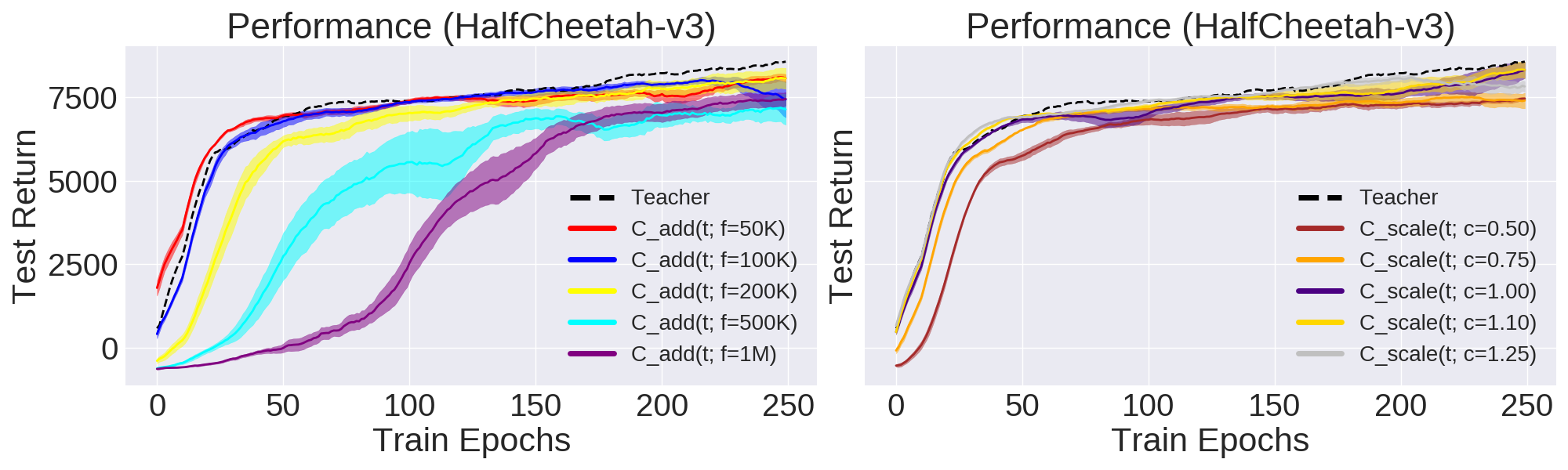}
\caption{
Offline student performance on HalfCheetah-v3 with additive curricula (left) and scale curricula (right); these experiments correspond to numerical values in Table~\ref{tab:results-250}. In both subplots, we show the teacher's performance for reference (at \emph{test} time, without noise) with dashed black lines. Results from additive curricula suggest a clear pattern that access to more samples initially (\ie larger $f$) slows learning.
}
\vspace*{-10pt}
\label{fig:results-250epochs-halfcheetah}
\end{figure*}

\section{Experiments}\label{sec:experiments}

\textbf{Environments}. We test \alg using standard MuJoCo environments~\cite{mujoco_2012} for Deep RL: Ant-v3, HalfCheetah-v3, Hopper-v3, and Walker2d-v3 from OpenAI~\cite{openaigym}. Earlier versions of these environments (-v1 or -v2) have also been benchmarked in work focusing on offline RL~\cite{BCQ_2019,BEAR_2019,behavior_regularized_2019}. 

\textbf{Teachers and Students}.
We train TD3 teachers using the standard 1M online steps~\cite{td3,sac,behavior_regularized_2019}, and store all encountered data tuples $(\bs, \ba, r, \bs')$ to make $\tbuf$. Each training epoch consists of 4000 time steps, so there are 250 training epochs for teachers and students. In Section~\ref{ssec:results-2500-epochs} we investigate training students for 10X more epochs.
We apply standard noise levels for the teacher's exploration; the noise added to actions is $\mathcal{N}(0, 0.1)$ instead of $\mathcal{N}(0, 0.5)$ as done in some experiments in~\cite{Optimistic_Perspective_2020,BCQ_2019} to increase data diversity.
In the Appendix, we have results from SAC~\cite{sac} teachers and students. The code we use is built on top of SpinningUp~\cite{SpinningUp2018}.

\textbf{Data Curriculum Experiments}.
We test a variety of additive and scale curricula from Section~\ref{ssec:curricula}. For additive curricula, we adjust the ``forward'' samples allowed: ${f \in \{50K, 100K, 200K, 500K, 1M\}}$, and we also check if ignoring older samples in $\tbuf$ helps (with $p=800K$). We report scale curricula with ${c \in \{0.50, 0.75, 1.00, 1.10, 1.25\}}$, where $c<1$ tests whether the student needs more gradient updates on data tuples in $\tbuf$ relative to the teacher, and ${c>1}$ tests whether it helps to have additional ``forward'' samples, which is similar to $f>0$ in additive curricula, but where $f$ increases throughout student training because the maximum eligible index $c\cdot t$ is a function of $t$.
Due to computational limitations, for most experiments after Section~\ref{ssec:results-250-epochs}, we test two particular curricula (from~\cite{BCQ_2019}): all the data ($C_{\rm add}(t; f=1M)$) or concurrent data ($C_{\rm scale}(t; c=1.00)$). 
We test with $C_{\rm add}(t; f=1M)$ because it serves a baseline of using all data, which can intuitively be viewed as using no curriculum.
As Deep RL evaluations are notoriously noisy~\cite{deep-rl-matters-2018}, all experiments are reproducible from code and data available on the project website.

\textbf{Evaluation}.
To evaluate student and teacher performance, we use two metrics, ``M1'' and ``M2'':
\begin{itemize}[leftmargin=*]
    \item \textbf{M1 (Final)}: average reward of the last 100 test episodes.
    \item \textbf{M2 (Average)}: average reward across all test episodes.
\end{itemize}
In all experiments, students and teachers do 10 test episodes after every epoch.
We use 5 random seeds for students, with respect to \emph{one} teacher (per environment),
to reduce the source of variability that would result from different teachers.
We compute M1 and M2 statistics for each of the 5 runs, then average those 5 to get final numbers for M1 and M2. In the tables, when comparing a relevant set of students, we bold the best M1 and M2 results, \emph{and additionally} bold other results with overlapping standard errors for a fairer comparison; see the Appendix for the exact formula.


\section{Results}\label{sec:results}

We present experimental results of \alg and report M1 and M2 metrics. We defer some results to the Appendix.

\subsection{Effect of Data Curricula on Student Training Offline}\label{ssec:results-250-epochs}

We train students offline using 11 data curricula and list results in Table~\ref{tab:results-250}.
The overall results suggest that a curriculum allowing for the available samples from $\tbuf$ to grow over time, and to include a few samples ``ahead'' relative to the student's training time $t$ produces stronger results. 
In particular, $C_{\rm scale}(t; c=1.10)$ has the best results, as it obtains the top scores or close to it (\ie with overlapping standard error) on 7 out of the 8 columns in Table~\ref{tab:results-250}.
The next best curriculum, with 4 out of 8 top scores, is $C_{\rm add}(t; f=50K)$, and it similarly allows the available range of samples to go slightly past $t$, in this case by a fixed $t+50K$.
In contrast, allowing too much data with $C_{\rm add}(t; f=1M)$ or $C_{\rm add}(t; f=500K)$ results in poor performance, with neither of these curricula ranking among the best in any of the environments with respect to either metric. 
Ignoring older samples from the teacher's early training history with $C_{\rm add}(t; p=800K, f=0)$ also exhibits weak performance.
See Figure~\ref{fig:results-250epochs-halfcheetah} for a representative set of learning curves for HalfCheetah-v3, showing some curricula that result in the student essentially matching teacher performance, despite known challenges associated with pure offline learning, even with relying on concurrent-style training~\cite{BCQ_2019}.

\begin{figure}[t]
\center
\includegraphics[width=0.50\textwidth]{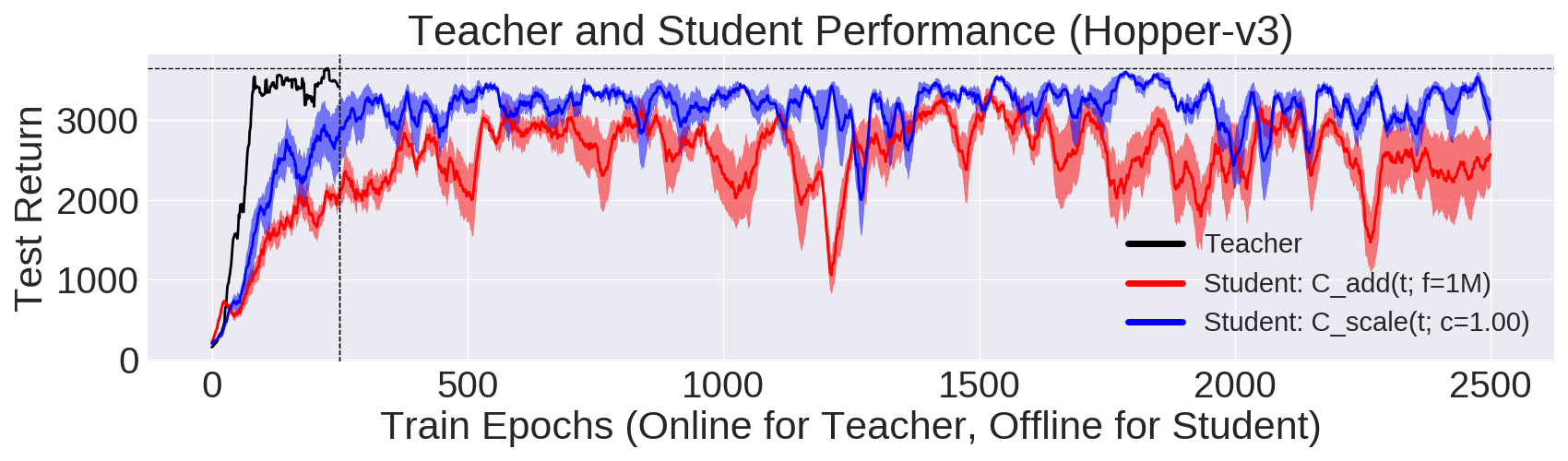}
\caption{
Hopper-v3 test-time episodic returns comparing the teacher performance (black curve) over its training period of 250 epochs (dashed vertical line), versus students trained offline over 2500 (10X more) epochs. For the two data curricula, we train 5 independent students from the fixed teacher data $\tbuf$ with different seeds. Results suggest that the $C_{\rm scale}(t; c=1.00)$ curriculum (blue curve) leads to better average performance over 2500 epochs, and matches the teacher's best performance (dashed horizontal line).
}
\vspace*{-10pt}
\label{fig:results-hopper-2500}
\end{figure}

\begin{table}[t]
 \setlength\tabcolsep{5.0pt}
 \caption{
 \textbf{Offline RL Results with \alg, 10X Training}. Student performance as a function of the data curriculum, where students train for 10X longer (2500 epochs) as compared to Table~\ref{tab:results-250}. Besides this change, the setup and table formatting is identical to Table~\ref{tab:results-250}. See Section~\ref{ssec:results-2500-epochs} for details. The teacher metrics are repeated for reference, and are not considered when bolding numbers here.
 }
 \begin{adjustbox}{width=\columnwidth,center}
 \centering
 \footnotesize
 \begin{tabular}{@{}lrrrrrrrr@{}} 
 \toprule
 & \multicolumn{2}{c}{Ant-v3} & \multicolumn{2}{c}{HalfCheetah-v3} & \multicolumn{2}{c}{Hopper-v3} & \multicolumn{2}{c}{Walker2d-v3}  \\
 \cmidrule(lr){2-3} \cmidrule(lr){4-5} \cmidrule(lr){6-7} \cmidrule(lr){8-9}
 Curriculum & M1 & M2 & M1 & M2 & M1 & M2 & M1 & M2 \\
 \midrule
 ${C_{\rm add}(t; \textrm{f=1M})}$     & \textbf{-2770.0} & -2088.3 & \textbf{6041.6} & 6847.9 & \textbf{2593.1} & 2482.0 & \textbf{3713.1} & 2763.5 \\
 ${C_{\rm scale}(t; \textrm{c=1.00})}$ & \textbf{-2780.6} & \textbf{-1511.2} & \textbf{6496.2} & \textbf{7555.5} & \textbf{3019.4} & \textbf{3052.3} & \textbf{3208.0} & \textbf{3121.4} \\
 \midrule
 TD3 Teacher & 4876.2 & 3975.8 & 8573.6 & 7285.5 & 3635.2 & 2791.9 & 3927.9 & 2579.8 \\
 \bottomrule \vspace{-0.5em} \\
 \end{tabular}
 \end{adjustbox}
 \label{tab:results-2500}
 \vspace{-1.0em}
\end{table}

\begin{table*}[t]
 \setlength\tabcolsep{5.0pt}
 \caption{
 \textbf{\alg Results with Online Data.} 
 Student performance based on one of two data curricula and the addition of a small fraction of online data. At the end of 250 training epochs, the student has collected self-generated, online data that constitutes 2.5\%, 5.0\%, or 10.0\% of the original teacher data size $|\tbuf|=1M$. We use the same teacher data as in Table~\ref{tab:results-250} and report the M1 and M2 metrics. We bold values by comparing only the two curricula with the same X\% online experiments per column and bolding the maximum only (if standard errors do not overlap) or both (if otherwise).
 }
 \centering
 \footnotesize
 \begin{tabular}{@{}lrrrrrrrr@{}} 
 \toprule
 & \multicolumn{2}{c}{Ant-v3} & \multicolumn{2}{c}{HalfCheetah-v3} & \multicolumn{2}{c}{Hopper-v3} & \multicolumn{2}{c}{Walker2d-v3}  \\
 \cmidrule(lr){2-3} \cmidrule(lr){4-5} \cmidrule(lr){6-7} \cmidrule(lr){8-9}
 Curriculum; \% Online & M1 & M2 & M1 & M2 & M1 & M2 & M1 & M2 \\
 \midrule
  ${C_{\rm add}(t ; f=1M)}$; 2.5\%     & 3093.1 & 894.4 & \textbf{8581.7} & 5275.3 & \textbf{3354.2} & 1977.9 & \textbf{3298.1} & 1882.6 \\
 ${C_{\rm scale}(t ; c=1.00)}$; 2.5\%  & \textbf{4004.7} & \textbf{2857.0} & \textbf{8417.9} & \textbf{7212.2} & 2712.9 & \textbf{2238.1} & \textbf{3144.9} & \textbf{2050.3} \\
 \midrule
 ${C_{\rm add}(t ; f=1M)}$; 5.0\%      & \textbf{3693.9} & 1556.6 & \textbf{8658.8} & 5634.6 & \textbf{3232.1} & 2230.5 & \textbf{3243.8} & \textbf{2097.3} \\
 ${C_{\rm scale}(t ; c=1.00)}$; 5.0\%  & \textbf{3251.1} & \textbf{3068.2} & \textbf{8601.4} & \textbf{7274.0} & \textbf{3356.2} & \textbf{2459.0} & \textbf{3155.8} & \textbf{2097.0} \\
 \midrule
 ${C_{\rm add}(t ; f=1M)}$; 10.0\%     & 4229.0 & 2007.6 & \textbf{8864.2} & 6024.5 & \textbf{2741.2} & 1979.4 & \textbf{3607.7} & \textbf{2448.2} \\
 ${C_{\rm scale}(t ; c=1.00)}$; 10.0\% & \textbf{4510.9} & \textbf{3349.0} & 8608.5 & \textbf{7290.5} & \textbf{3182.9} & \textbf{2580.6} & 3146.9 & 2204.2 \\
 \midrule
 TD3 Teacher                           & 4876.2 & 3975.8 & 8573.6 & 7285.5 & 3635.2 & 2791.9 & 3927.9 & 2579.8 \\
 \bottomrule \vspace{-0.5em} \\
 \end{tabular}
 \label{tab:results-online}
 \vspace{-1.0em}
\end{table*}

\begin{figure*}[h]
\center
\includegraphics[width=0.81\textwidth]{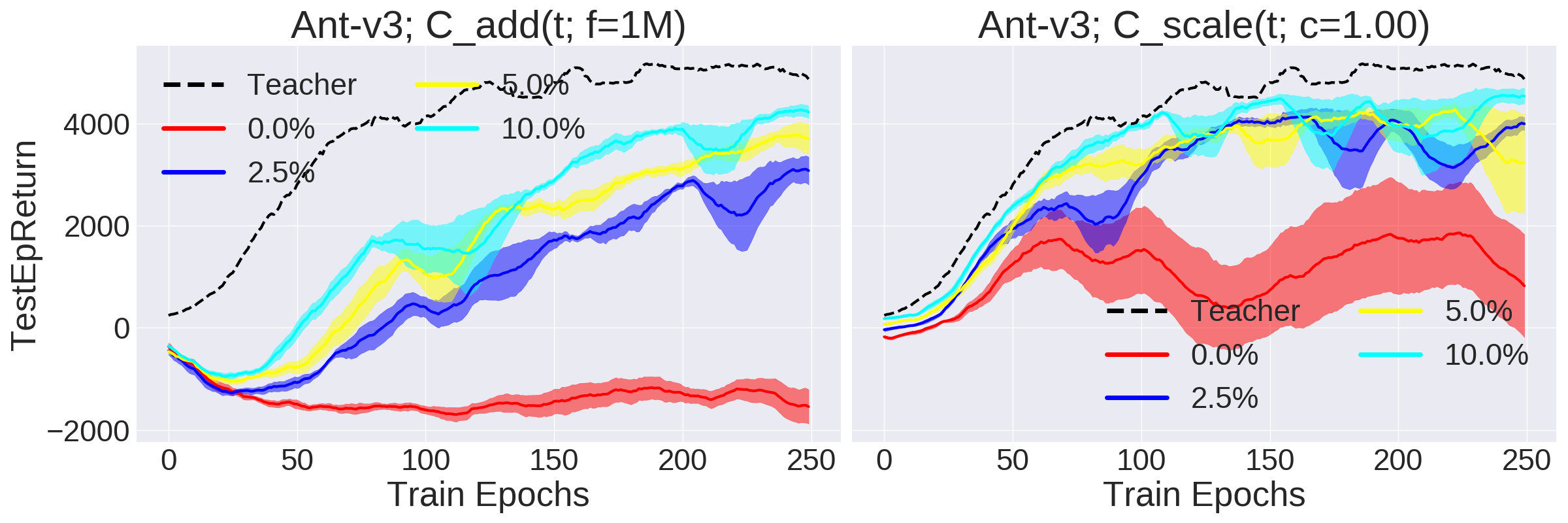}
\caption{
Ant-v3 student test performance with various amounts of online data, from 0\% (\ie offline) to 10\% online. We show the teacher curve (dashed black line) for reference. We plot results from the two curricula tested in Table~\ref{tab:results-online}, $C_{\rm add}(t; f=1M)$ (left), and $C_{\rm scale}(t; c=1.00)$ (right). 
}
\vspace*{-5pt}
\label{fig:online-ant-comparisons}
\end{figure*}

\begin{figure*}[t]
\center
\includegraphics[width=0.95\textwidth]{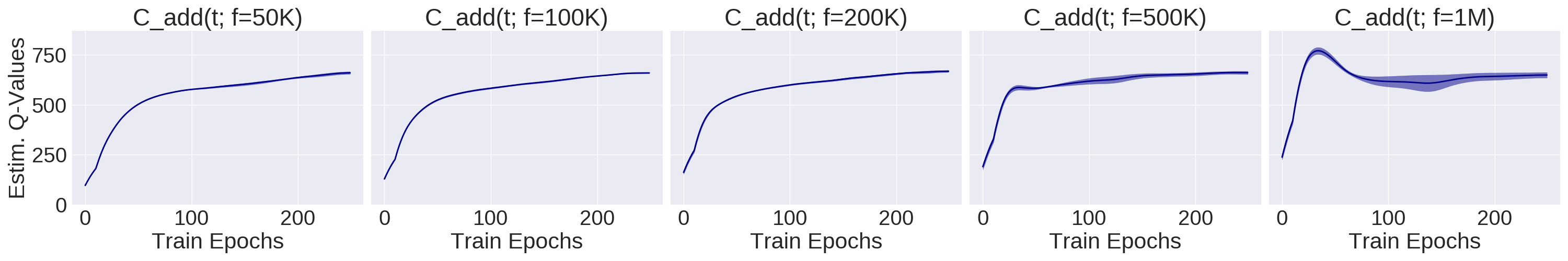}
\caption{
Students' estimated Q-values in HalfCheetah-v3.
With more data available at the start (\ie increasing $f$, shown left to right) this creates initial over-estimation of Q-values. 
All 5 subplots are derived from students reported in Table~\ref{tab:results-250}, and all curves average over the 5 seeds and show standard errors.
}
\vspace*{-10pt}
\label{fig:results-qvalues-hc}
\end{figure*}

\subsection{Data Curriculum for Training 10X Longer}\label{ssec:results-2500-epochs}

Here, as in Section~\ref{ssec:results-250-epochs}, we run the ${C_{\rm add}(t ; f=1M)}$ and ${C_{\rm scale}(t ; c=1.00)}$ curricula again, but train 10X longer to understand how this affects learning.
The choice of curriculum will \emph{only} affect the first 1/10 of training (\ie the first 250 epochs), since after that, both curricula reduce to the student sampling data tuples from the entire teacher buffer $\tbuf$. To make results comparable with those in Section~\ref{ssec:results-250-epochs}, we use the same teacher buffers.
Table~\ref{tab:results-2500} has results in a similar manner as in Table~\ref{tab:results-250}. We find that, somewhat surprisingly, the curricula affects the long-term performance of the student in the remaining 9/10 of training. Across all 8 columns (4 environments and the M1/M2 metrics), using ${C_{\rm scale}(t ; c=1.00)}$ outperforms ${C_{\rm add}(t; f=1M)}$ or is statistically similar to it based on the standard error metric we use (see Section~\ref{sec:experiments}), suggesting that the initial curriculum assists TD3 in finding a stable set of policy and value functions so that it can continue training without significant deterioration.
Figure~\ref{fig:results-hopper-2500} shows a representative set of learning curves for Hopper-v3, which shows the student with $C_{\rm scale}(t; c=1.00)$ matching the teacher's performance given sufficient training.

\subsection{Results with Small Amounts of Online Data}\label{ssec:apprenticeship-results}

We next study when students can use a small amount of online data to address challenges with pure offline learning.
We train students for 250 epochs using 2.0\%, 5.0\% and 10.0\% online data collection, \ie at the \emph{end} of 1M training steps, students get 25K, 50K, and 100K self-generated data tuples, respectively, for $\sbuf$, which they can sample from in addition to the (filtered) data tuples from $\tbuf$.
Table~\ref{tab:results-online} has a complete overview of the M1 and M2 results across all 4 environments, and the Appendix contains further details.
The results indicate that even with as little as 2.5\% online data, students are able to significantly improve versus offline learning, with the improvement most notable in the complex Ant-v3 environment as shown in Figure~\ref{fig:online-ant-comparisons}.
Furthermore, $C_{\rm scale}(t; c=1.00)$ continues to provide some benefit to learning speed compared to $C_{\rm add}(t; f=1M)$ with online data, particularly with respect to the M2 metric.

\subsection{Investigation of Q-Values}\label{ssec:q-values}

To understand why the data curriculum matters, we investigate the student's estimated Q-values. With too much data available from $\tbuf$ during early stages of training, this causes overestimation of Q-values, whereas a curriculum that restricts data tuples results in more stable, monotonically increasing Q-values. Figure~\ref{fig:results-qvalues-hc} shows different additive curricula on HalfCheetah-v3 and plots the estimated Q-values, where the pattern of an initial ``hump'' at the start is prominent for certain curricula, particularly $C_{\rm add}(t; f=1M)$. As shown in the Appendix, a similar trend holds for other environments.


\section{Conclusion and Future Work}\label{sec:conclusion}


This work introduces the \alg framework, which studies how to filter a given dataset for existing RL algorithms. Results across a variety of curricula and training settings suggest that the choice significantly impacts the learning speed of students running RL.
The greatest benefits come from curricula that gradually let the available data grow as a function of training time.
We caution that the results presented are contingent on the way we generated the datasets.
In future work, we will test other datasets and environments, such as D4RL~\cite{datasets_offline_rl_2020} or RL Unplugged~\cite{RLUnplugged_2020}, possibly with multiple teachers~\cite{acteach}. While we kept relevant experience replay hyperparameters such as the \emph{replay ratio} constant, we will use findings from recent research~\cite{Revisiting_XP_Replay_2020} to investigate the interplay between experience replay and data curricula. Finally, we plan to devise more sophisticated data curricula.


\section*{Acknowledgments}

{\footnotesize Daniel Seita was supported by the GFSD throughout this research. We thank members of the CannyLab for helpful advice and suggestions.}


\clearpage 
\bibliographystyle{IEEEtranS}
\bibliography{example}

\normalsize
\cleardoublepage
\appendices

\begin{center}
\Large \textbf{Supplementary Material}
\end{center}

In this supplementary material, we
\begin{itemize}
    \item discuss the environments and teacher policy performance (Section~\ref{app:teacher-performance}),
    \item describe additional experiment details (Section~\ref{app:experiment-details}),
    \item present additional results (Section~\ref{app:experiment-results}), including new experiments that test SAC teachers and students (Section~\ref{ssec:sac-results}) and \emph{cross-algorithm teaching} where we test SAC teachers and TD3 students, and vice versa (Section~\ref{app:cross-alg-teaching}),
    \item list ideas we tried but which did not yield strong results (Section~\ref{app:did-not-work}).
\end{itemize}

Additional material is available at: {\small \url{https://tinyurl.com/teach-dcur}}.


\section{Environments and Teacher Performance}\label{app:teacher-performance}

\begin{figure*}[h]
\center
\includegraphics[width=1.00\textwidth]{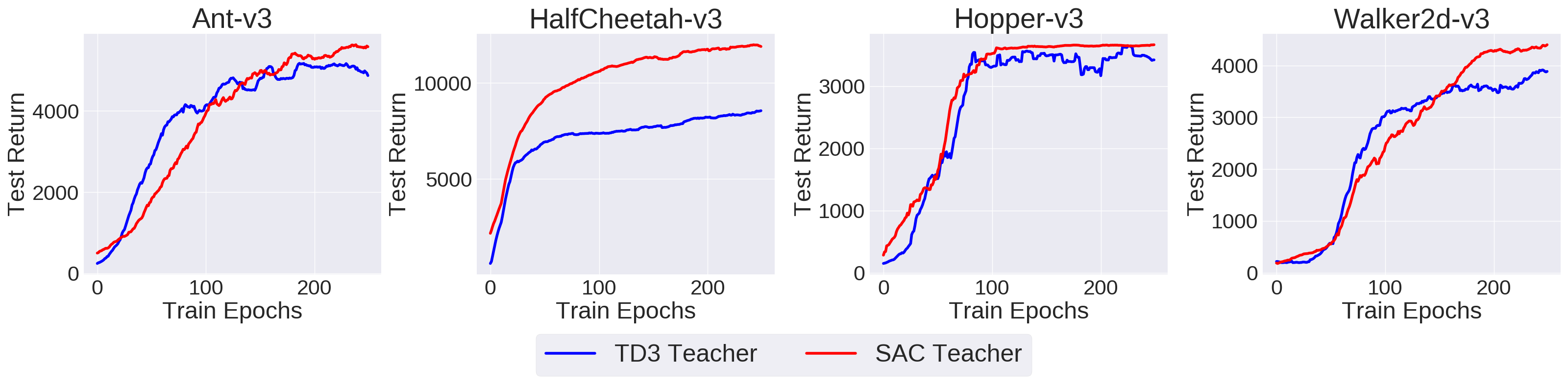}
\caption{
Teacher performance of TD3 (blue) and SAC (red) on the four environments we test in this work. 
}
\vspace*{-5pt}
\label{fig:teacher-performance}
\end{figure*}

We use the -v3 version of the MuJoCo environments from OpenAI gym~\cite{openaigym}, which is the latest version available at the time of this research. We also use version 2.00 for MuJoCo itself. The state $\bs$ and action $\ba$ dimensions for the environments are:
\begin{itemize}[noitemsep]
\item \textbf{Ant-v3}: state 111, action 8.
\item \textbf{HalfCheetah-v3}: state 17, action 6.
\item \textbf{Hopper-v3}: state 11, action 3.
\item \textbf{Walker2d-v3}: state 17, action 5.
\end{itemize}
We build on top of the SpinningUp code~\cite{SpinningUp2018} in this work to generate TD3 and SAC teachers (and students). For both algorithms, we use seed 40 to train HalfCheetah-v3 and Hopper-v3 teachers, and seed 50 to train Ant-v3 and Walker2d-v3 teachers (due to seed 40 resulting in poor TD3 performance in these environments), and save the resulting 8 data buffers to form the $\tbuf$ datasets that students learn from in subsequent experiments.

Figure~\ref{fig:teacher-performance} shows the TD3 and SAC teacher performance.
The curves show the test time performance after each epoch by rolling out the (deterministic) policies for 10 episodes and averaging the resulting episodic rewards. Then we smooth the curves with a 20-window moving average.
These results are comparable with prior results in the research literature~\cite{sac,td3}, though we caution that these works use -v1 of the MuJoCo environments, which may induce subtle differences.


\section{Additional Experiment Details}\label{app:experiment-details}

For teachers and students, we use the default hyperparameters from SpiningUp for TD3 and SAC to ensure minimal changes from standard Deep RL implementations, as small adjustments can lead to significant performance differences~\cite{deep-rl-matters-2018}.
The main exception is that we increase the default number of epochs from 100 to 250 to get 1M time steps.\footnote{For the longer-horizon experiments from Section~\ref{ssec:results-2500-epochs}, we increase the epoch count to 2500.}
The policy $\pi$ and value $Q$ networks are fully connected with two hidden layers, each of which have 256 nodes. The policy network outputs a vector in $\mathbb{R}^n$ where $n$ is the action dimension, and the value function takes in the concatenated state and action $(\bs,\ba)$ and output a single scalar in $\mathbb{R}$.

We note a subtle point regarding training and time steps: as mentioned in Section~\ref{sec:ps}, we use the term ``time step'' to refer to both an online environment step and a gradient update, since in MuJoCo environments, the ratio between the two tends to be one~\cite{SpinningUp2018}. 
However, this ignores a slight subtlety with the initial period of training, where it is common to perform several thousand environment interactions with a purely random policy to sufficiently populate a replay buffer before conducting gradient updates. The default in SpinningUp is to perform 1000 random steps,\footnote{In SpinningUp, this is the \texttt{update\_after} hyperparameter for TD3 and SAC.} which we use in this work. Thus, to make the students have the same number of gradient updates as a standard online RL agent, we start the student ``time step'' $t$ at $t=1000$. The students do one gradient update per time step up until $t=1M$, assuming a standard run of 250 epochs. In this case, the first epoch will have 3000 gradient updates, and all the remaining 249 epochs have 4000 gradient updates.

\subsection{Evaluation Performance with Standard Errors}\label{app:standard_error_overlap}

We use a fixed set of random seeds (90 through 94) for students in all experiments, so any presented student results are aggregates over 5 independent learning runs initialized at those random seeds. Due to the limited number of trials, we do not report confidence intervals or bootstrapped estimates. Instead, we rely on the standard error of the mean, which is the sample standard deviation divided by $\sqrt{5}$. 
For tables in the main part of the paper, we omit the standard errors to aid readability of student results, and instead report the standard errors throughout Section~\ref{app:experiment-results}. 

For reporting results in tables, we compare among the effects of different data curriculum choices. The convention is to bold the best student M1 and M2 performance, which are separated by columns. \emph{In addition}, we bold any other statistic which has overlapping standard errors as a fairer way of comparing different data curricula due to the highly stochastic nature of Deep RL~\cite{deep-rl-matters-2018}.
More formally, consider a data curriculum for the student that results in the statistic for M1 (without loss of generality, M2) with the highest value, which is $w \pm x$ for standard error $x$. Then, for any other student data curriculum statistic, we have some performance $y \pm z$ with standard error $z$. By assumption, $w > y$. We do \emph{not} bold face the result $y$ if the following inequality holds:
\begin{equation}\label{ineq:bold}
w - x > y + z,
\end{equation}
which suggests that even when subtracting the standard error from $w$ and adding the standard error to $y$, the choice of data curriculum that led to the result of $w$ still outperforms the one that produced $y$. We bold face the result $y$ in tables if the inequality~(\ref{ineq:bold}) does \emph{not} hold, since that suggests the difference between the two statistics may not be significant.

\subsection{Overlap Analysis}\label{ssec:overlap}

To benchmark learning progress or to select appropriate curricula or teachers for a given student, we propose to measure the ``similarity'' in the distribution of states $\bs$ induced by the corresponding policies $\pi_{\theta_S}$ and $\pi_{\theta_T}$.
We study this relationship using a novel \emph{overlap} function, which may be of independent interest outside of \alg. For students and teachers, we denote $\mathcal{D}_1$ and $\mathcal{D}_2$ as equal-sized sets of states induced by their respective policies. We measure overlap $f_{\rm olap}(\mathcal{D}_1, \mathcal{D}_2)$ by training a binary classifier $g_\phi(\bs) \to [0,1]$ to classify states $\bs$ as belonging to $\mathcal{D}_1$ or $\mathcal{D}_2$, and computing:
\begin{equation}\label{eq:overlap}
    f_{\rm olap}(\mathcal{D}_1, \mathcal{D}_2) = 2 \cdot (1 - {\rm Acc})
\end{equation}
where ${\rm Acc}$ is validation accuracy after sufficient training (\ie when it stops improving). Hence, validation accuracy of 1 means $f_{\rm olap}=0$, implying that there is maximum ``separation'' between the state distribution induced from $\pi_{\theta_S}$ and $\pi_{\theta_T}$. At the other extreme, accuracy of 0.5 means $g_\phi$ has the same performance as random guessing, hence $f_{\rm olap}=1$ with maximum overlap, suggesting that the policies $\pi_{\theta_S}$ and $\pi_{\theta_T}$ are indistinguishable.

After every student training epoch, we execute its current noise-free policy to get 50K states $\bs \in \mathcal{S}$ to form $\mathcal{D}_1$. We compare this with an equal number of states subsampled from the teacher data $\tbuf$ \emph{after} applying the curriculum (Section~\ref{ssec:curricula}), which forms $\mathcal{D}_2$. If the curriculum enables fewer than 50K teacher states at the current student training epoch, we use all the available teacher states and subsample from the 50K student states to ensure the two classes have equal sample counts. The overlap network $g_\phi(\bs)$ follows the same architecture as the policy networks except the final output is a single scalar. We do not test with ``state-action overlap'' and leave this analysis to future work.


\section{Additional Experiment Results}\label{app:experiment-results}

We report additional results and statistics that we could not include earlier. Sections~\ref{app:250-epochs},~\ref{app:2500-epochs},~\ref{app:more-apprenticeship-results}, and~\ref{app:q-values} expand upon sections from the main part of the paper. 
Sections~\ref{ssec:sac-results} and~\ref{app:cross-alg-teaching} report brand new sets of experiments, respectively, for SAC teachers and students, and for cross-algorithm teaching. In the latter, we test with SAC teachers and TD3 students and vice versa, along with TD3 teachers and BCQ~\cite{BCQ_2019} students.

In tables here, we continue reporting M1 and M2 statistics (Section~\ref{sec:experiments}) and also report standard errors, \ie the sample standard deviation divided by $\sqrt{5}$. See Section~\ref{app:standard_error_overlap} for a further discussion of standard errors.
In all figures in this paper with student learning curves or other student statistics (\eg estimated Q-values), we shade in the standard error of the mean and smooth the curves over a moving average window of length 20 to improve legibility.

\subsection{More Detailed Results from Section~\ref{ssec:results-250-epochs}}\label{app:250-epochs}

\begin{figure*}[t]
\center
\includegraphics[width=0.90\textwidth]{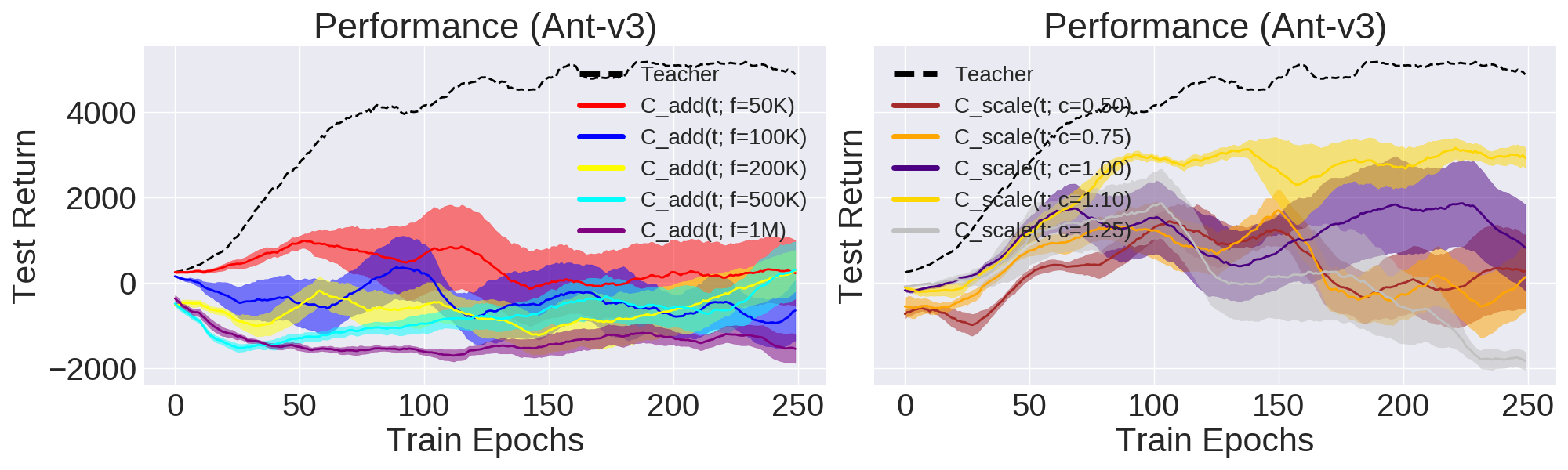}
\caption{
Offline TD3 student performance on Ant-v3 based on additive curricula (left) and scale curricula (right), shown in the same way as Figure~\ref{fig:results-250epochs-halfcheetah}.
}
\vspace*{-5pt}
\label{fig:results-250epochs-ant}
\end{figure*}

\begin{figure*}[t]
\center
\includegraphics[width=0.90\textwidth]{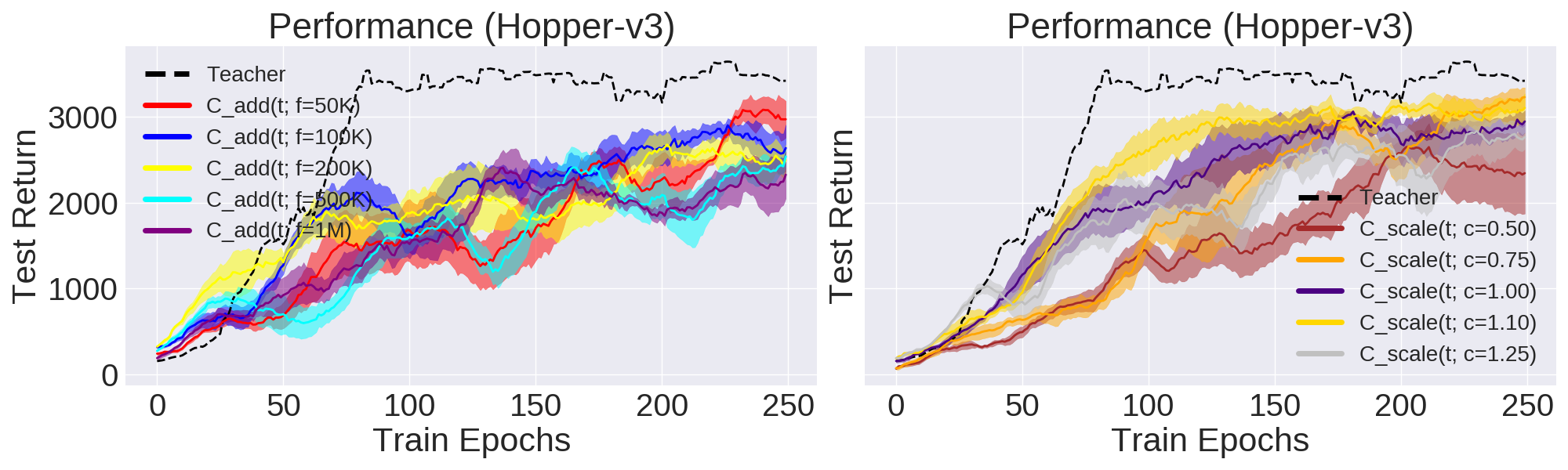}
\caption{
Offline TD3 student performance on Hopper-v3 based on additive curricula (left) and scale curricula (right), shown in the same way as Figure~\ref{fig:results-250epochs-halfcheetah}.
}
\vspace*{-5pt}
\label{fig:results-250epochs-hopper}
\end{figure*}

\begin{figure*}[t]
\center
\includegraphics[width=0.90\textwidth]{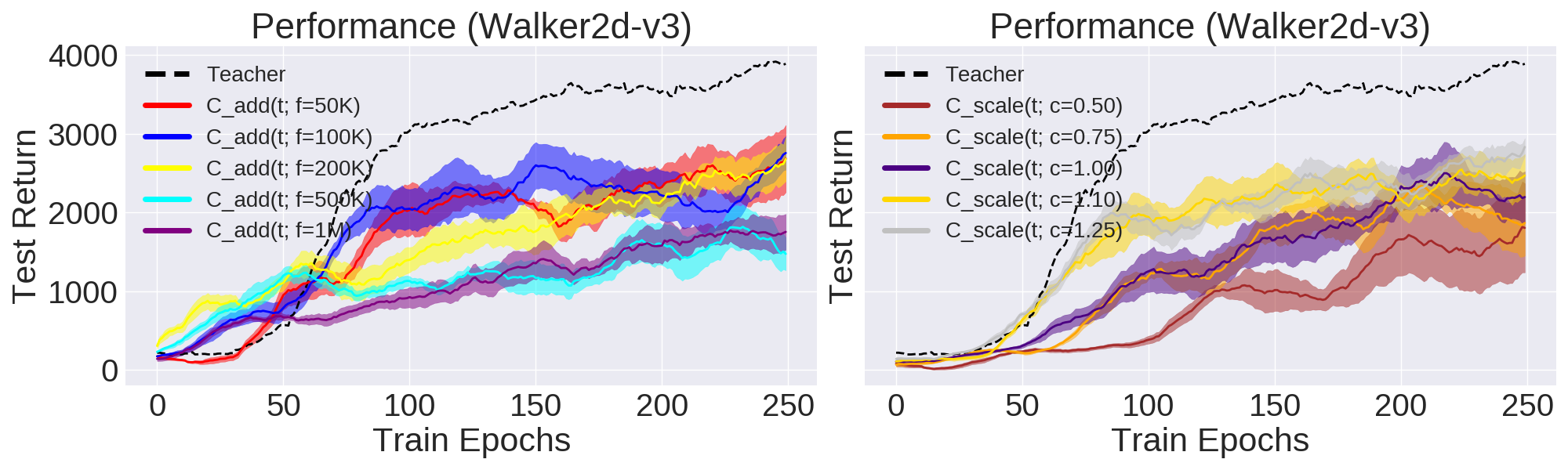}
\caption{
Offline TD3 student performance on Walker2d-v3 based on additive curricula (left) and scale curricula (right), shown in the same way as Figure~\ref{fig:results-250epochs-halfcheetah}.
}
\vspace*{-5pt}
\label{fig:results-250epochs-walker2d}
\end{figure*}

\begin{table*}[t]
 \setlength\tabcolsep{5.0pt}
 \caption{
 \textbf{Offline RL Results with \alg}. M1 and M2 results for students (and teacher in the bottom row), reported in a similar manner as in Table~\ref{tab:results-250} but with standard error included.
 }
 \centering
 \footnotesize
 \begin{tabular}{@{}LRRRRRRRR@{}} 
 \toprule
 & \multicolumn{2}{c}{Ant-v3} & \multicolumn{2}{c}{HalfCheetah-v3} & \multicolumn{2}{c}{Hopper-v3} & \multicolumn{2}{c}{Walker2d-v3} \\
 \cmidrule(lr){2-3} \cmidrule(lr){4-5} \cmidrule(lr){6-7} \cmidrule(lr){8-9}
 Data Curriculum & M1 & M2 & M1 & M2 & M1 & M2 & M1 & M2 \\
 \midrule
 \rowfont{\footnotesize} 
 ${C_{\rm add}(t ; p=1M,f=50K)}$  &   219.9 & 382.0 & \textbf{8067.3} & \textbf{7147.0}  & 2986.2 & 1669.5 & \textbf{2715.6} & \textbf{1712.1}  \\
 \rowfont{\scriptsize} 
                                  &   $\pm$ 765.0  & $\pm$ 504.3  & \textbf{$\pm$ 123.6} & \textbf{$\pm$ 45.7} & $\pm$ 191.3 & $\pm$ 141.3  & \textbf{$\pm$ 404.1} & \textbf{$\pm$ 148.6}   \\
 \rowfont{\footnotesize} 
 ${C_{\rm add}(t ; p=1M,f=100K)}$ &  -604.2 & -409.6 & 7416.2 & \textbf{7052.6} & 2627.4 & 1980.1  & \textbf{2746.7} & \textbf{1810.5}   \\
 \rowfont{\scriptsize} 
                                  &  $\pm$ 724.9  & $\pm$ 420.5 & $\pm$ 540.2 & \textbf{$\pm$ 83.6}  & $\pm$ 272.3 & $\pm$ 64.1   & \textbf{$\pm$ 231.3} & \textbf{$\pm$ 191.5}  \\
 \rowfont{\footnotesize} 
 ${C_{\rm add}(t ; p=1M,f=200K)}$ &   288.4 & -620.4  & \textbf{8028.4} & 6521.5  & 2525.3  & 1887.9   & \textbf{2691.8} & \textbf{1651.9} \\
 \rowfont{\scriptsize} 
                                  &  $\pm$ 454.4  & $\pm$ 385.4 & \textbf{$\pm$ 321.1} & $\pm$ 138.7 & $\pm$ 173.0 & $\pm$ 99.1   & \textbf{$\pm$ 266.4} & \textbf{$\pm$  38.3}  \\
 \rowfont{\footnotesize} 
 ${C_{\rm add}(t ; p=1M,f=500K)}$ &   283.3  & -801.1  & 7108.7 & 5041.2  & 2498.0 & 1546.2  & 1515.7 & 1183.5 \\
 \rowfont{\scriptsize} 
                                  &  $\pm$ 667.1  & $\pm$ 254.5 & $\pm$ 449.6 & $\pm$ 307.2 & $\pm$ 105.5 & $\pm$ 98.1   & $\pm$ 214.4 & $\pm$  81.8  \\
 \rowfont{\footnotesize} 
 ${C_{\rm add}(t ; p=1M,f=1M)}$   & -1552.1  & -1390.5 & 7447.3 & 3846.8  & 2323.0 & 1610.3  & 1741.4  & 1096.4  \\
 \rowfont{\scriptsize} 
                                  &  $\pm$ 354.0  & $\pm$ 90.0 & $\pm$ 187.8 & $\pm$ 237.9 & $\pm$ 323.0 & $\pm$ 77.3   & $\pm$ 226.8 & $\pm$  76.9  \\
 \rowfont{\footnotesize} 
 ${C_{\rm add}(t ; p=800K,f=0)}$  &  -889.7  & 1497.9 & \textbf{7650.4} & 6942.7 &  2132.2 & 1924.4  &  792.0 & 1521.8 \\
 \rowfont{\scriptsize} 
                                  &  $\pm$ 516.1  & $\pm$ 420.5 & \textbf{$\pm$ 806.4} & $\pm$ 142.0  & $\pm$ 255.2 & $\pm$ 132.5  &  $\pm$ 343.2 & $\pm$  74.7  \\
 \rowfont{\footnotesize} 
 ${C_{\rm scale}(t ; c=0.50)}$    &   285.0  & 301.4  & 7467.8  & 6261.9  & 2332.3 & 1450.5  & 1866.0 &  799.9  \\
 \rowfont{\scriptsize} 
                                  &  $\pm$ 843.9 & $\pm$ 318.9  & $\pm$ 52.3  & $\pm$ 83.9  & $\pm$ 467.5 & $\pm$ 86.2  & $\pm$ 589.1 &  $\pm$ 111.5  \\
 \rowfont{\footnotesize} 
 ${C_{\rm scale}(t ; c=0.75)}$    &   167.0  & 412.2 & 7392.6 & 6689.2  & \textbf{3284.7} & 1818.2  & 1864.9  & 1251.5  \\
 \rowfont{\scriptsize} 
                                  &  $\pm$  791.9 & $\pm$ 179.0  & $\pm$ 235.4 & $\pm$ 67.5  & \textbf{$\pm$ 103.0} & $\pm$ 101.9  & $\pm$ 410.8 & $\pm$  71.4  \\
 \rowfont{\footnotesize} 
 ${C_{\rm scale}(t ; c=1.00)}$    &   825.6  & 1103.7 & \textbf{8305.3} & 6980.0 & 2984.3  & 2092.6  & 2178.5 & 1305.7  \\
 \rowfont{\scriptsize} 
                                  &  $\pm$ 1022.4 & $\pm$ 628.9 & \textbf{$\pm$ 246.2} & $\pm$ 111.6 & $\pm$ 170.3 & $\pm$ 84.8   & $\pm$ 317.5 & $\pm$  86.6 \\
 \rowfont{\footnotesize} 
 ${C_{\rm scale}(t ; c=1.10)}$    &  \textbf{2952.5} & \textbf{2212.0} & \textbf{8306.0} & \textbf{7095.6}  & \textbf{3185.2} & \textbf{2317.1}   & 2423.9  & \textbf{1698.1}  \\
 \rowfont{\scriptsize} 
                                  &  \textbf{$\pm$  217.5} & \textbf{$\pm$ 194.4} & \textbf{$\pm$ 255.1} & \textbf{$\pm$ 79.4}  & \textbf{$\pm$ 174.4} & \textbf{$\pm$ 60.1}   & $\pm$ 278.7 & \textbf{$\pm$  59.6}  \\
 \rowfont{\footnotesize} 
 ${C_{\rm scale}(t ; c=1.25)}$    & -1851.1  & 199.6  & 7843.8 & \textbf{7175.4}  & 2755.5 & 1891.9  & \textbf{2839.4} & \textbf{1747.4}  \\
 \rowfont{\scriptsize} 
                                  &  $\pm$  197.1 & $\pm$ 515.2  & $\pm$ 150.2 & \textbf{$\pm$ 39.4} & $\pm$ 113.8 & $\pm$ 107.0  & \textbf{$\pm$ 126.6} & \textbf{$\pm$  85.8}   \\
 \midrule
 \rowfont{\footnotesize} 
 TD3 Teacher     & 4876.2 & 3975.8 & 8573.6 & 7285.5 & 3635.2 & 2791.9 & 3927.9 & 2579.8 \\
 \bottomrule \vspace{-0.5em} \\
 \end{tabular}
 \label{tab:results-250-ant-hc}
 \vspace{-1.0em}
\end{table*}

We show additional learning curves for TD3 students learning entirely offline with different curricula, corresponding to experiments from Section~\ref{ssec:results-250-epochs}.
Figures~\ref{fig:results-250epochs-ant},~\ref{fig:results-250epochs-halfcheetah},~\ref{fig:results-250epochs-hopper}, and~\ref{fig:results-250epochs-walker2d} show performance across 250 epochs for Ant-v3, HalfCheetah-v3, Hopper-v3, and Walker2d-v3, respectively. 
In Table~\ref{tab:results-250-ant-hc}, we show the corresponding M1 and M2 statistics for all four environments. These are the same numbers in Table~\ref{tab:results-250} except with standard errors now included.

\subsection{More Detailed Results from Section~\ref{ssec:results-2500-epochs}}\label{app:2500-epochs}

\begin{figure*}[t]
\center
\includegraphics[width=0.75\textwidth]{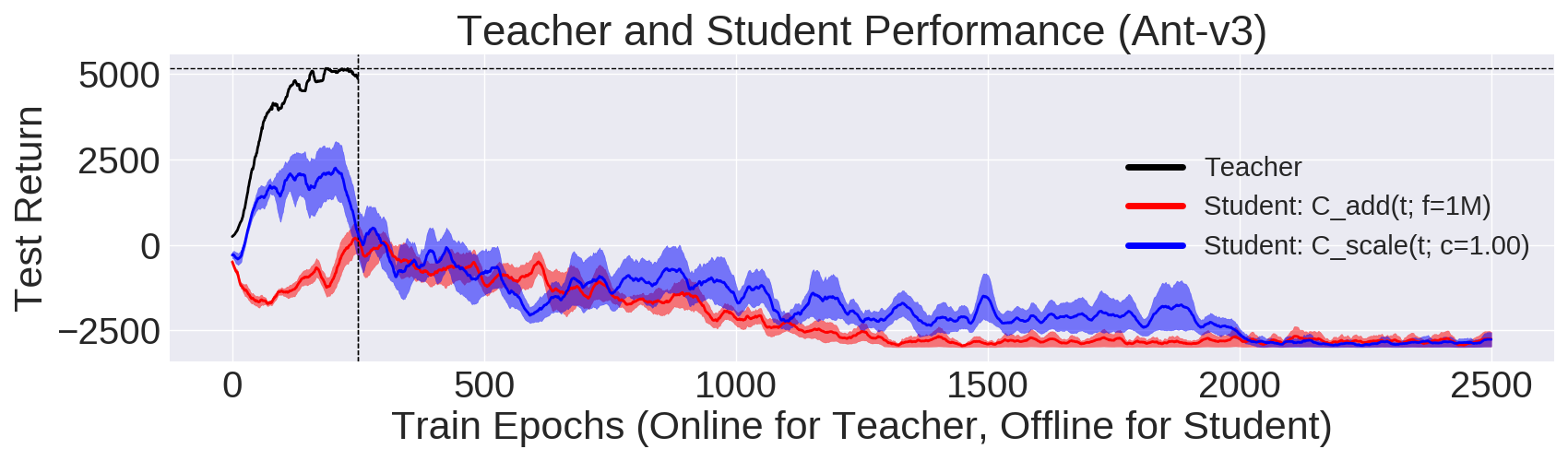}
\caption{
Results for Ant-v3, presented in a similar manner as Figure~\ref{fig:results-hopper-2500}. Results suggest difficulties in learning from offline data alone.
}
\vspace*{-5pt}
\label{fig:results-ant-2500}
\end{figure*}

\begin{figure*}[t]
\center
\includegraphics[width=0.75\textwidth]{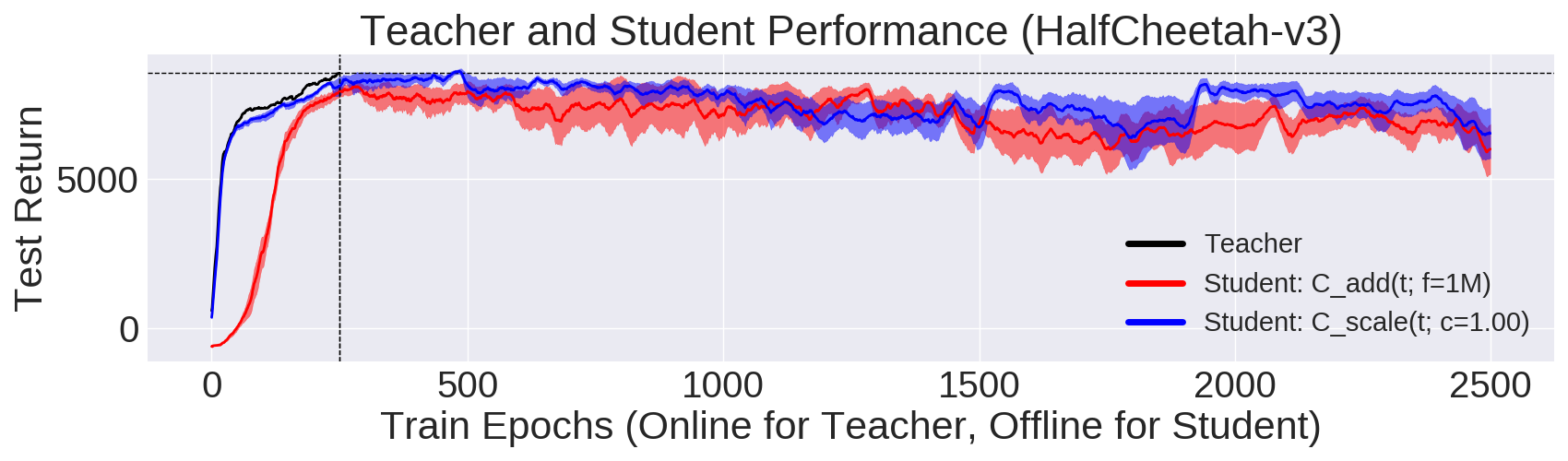}
\caption{
Results for HalfCheetah-v3, presented in a similar manner as Figure~\ref{fig:results-hopper-2500}. Both curriculum choices have stable training, with similar performance after 250 epochs.
}
\vspace*{-5pt}
\label{fig:results-halfcheetah-2500}
\end{figure*}

\begin{figure*}[t]
\center
\includegraphics[width=0.75\textwidth]{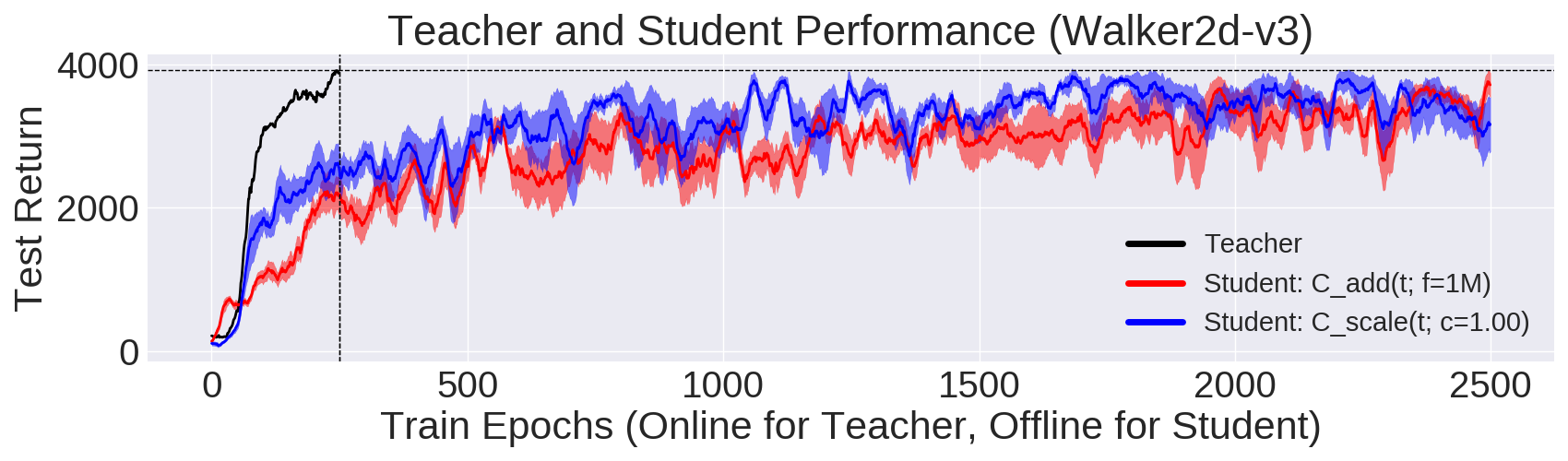}
\caption{
Results for Walker2d-v3, presented in a similar manner as Figure~\ref{fig:results-hopper-2500}. Both curriculum choices have stable training, with $C_{\rm scale}(t; c=1.00)$ showing slightly better performance across training.
}
\vspace*{-5pt}
\label{fig:results-walker2d-2500}
\end{figure*}

\begin{table*}[t]
 \setlength\tabcolsep{5.0pt}
 \caption{
 \textbf{\alg Results, 10X Offline Training}. Results from Table~\ref{tab:results-2500}, but with standard errors. See Section~\ref{app:2500-epochs} for further details.
 }
 \centering
 \footnotesize
 \begin{tabular}{@{}LRRRRRRRR@{}} 
 \toprule
 & \multicolumn{2}{c}{Ant-v3} & \multicolumn{2}{c}{HalfCheetah-v3} & \multicolumn{2}{c}{Hopper-v3} & \multicolumn{2}{c}{Walker2d-v3}  \\
 \cmidrule(lr){2-3} \cmidrule(lr){4-5} \cmidrule(lr){6-7} \cmidrule(lr){8-9}
 Curriculum & M1 & M2 & M1 & M2 & M1 & M2 & M1 & M2 \\
 \midrule
 \rowfont{\footnotesize} 
    ${C_{\rm add}(t ; f=1M)}$     & \textbf{-2770.0} & -2088.3 & \textbf{6041.6} & 6847.9  & \textbf{2593.1} & 2482.0 & \textbf{3713.1} & 2763.5  \\
 \rowfont{\scriptsize}
    & \textbf{$\pm$ 205.5} & $\pm$ 141.9 & \textbf{$\pm$ 857.7} & $\pm$ 348.3  & \textbf{$\pm$ 386.6} & $\pm$ 61.5 & \textbf{$\pm$ 163.6} & $\pm$ 131.9 \\
 \rowfont{\footnotesize} 
    ${C_{\rm scale}(t ; c=1.00)}$ & \textbf{-2780.6} & \textbf{-1511.2} & \textbf{6496.2} & \textbf{7555.5}  & \textbf{3019.4} & \textbf{3052.3} & \textbf{3208.0} & \textbf{3121.4}  \\
 \rowfont{\scriptsize}
    & \textbf{$\pm$ 195.1} & \textbf{$\pm$ 225.2} & \textbf{$\pm$ 835.8} & \textbf{$\pm$ 230.2} & \textbf{$\pm$ 227.1} & \textbf{$\pm$ 49.7} & \textbf{$\pm$ 367.4} & \textbf{$\pm$ 92.7}  \\
 \midrule
 \rowfont{\footnotesize} 
    TD3 Teacher & 4876.2 & 3975.8 & 8573.6 & 7285.5 & 3635.2 & 2791.9 & 3927.9 & 2579.8 \\
 \bottomrule \vspace{-0.5em} \\
 \end{tabular}
 \label{tab:results-2500-stderr-1}
 \vspace{-1.0em}
\end{table*}

This section expands upon Section~\ref{ssec:results-2500-epochs} where students run 10X more gradient steps than usual.
Figures~\ref{fig:results-ant-2500},~\ref{fig:results-halfcheetah-2500},~\ref{fig:results-hopper-2500}, and~\ref{fig:results-walker2d-2500} show the performance for Ant-v3, HalfCheetah-v3, Hopper-v3, and Walker2d-v3, respectively. The corresponding numerical metrics M1 and M2 are reported in Table~\ref{tab:results-2500-stderr-1}.
The results show that for 3 of the 4 environments, training for longer on the fixed teacher data $\tbuf$ (possibly with the $C_{\rm scale}(t; c=1.00)$ curriculum) can let the student attain performance which essentially matches that of the best teacher performance, and furthermore, that there is little deterioration despite the difficulties with offline learning.
The exception is Ant-v3, which is the most challenging of these environments.

\subsection{More Detailed Results from Section~\ref{ssec:apprenticeship-results}}\label{app:more-apprenticeship-results}

\begin{table*}[t]
 \setlength\tabcolsep{5.0pt}
 \caption{
 \textbf{\alg Results with Online Data.} Results from Table~\ref{tab:results-online}, but with standard errors. See Section~\ref{app:more-apprenticeship-results} for further details.
 }
 \centering
 \footnotesize
 \begin{tabular}{@{}LRRRRRRRR@{}} 
 \toprule
 & \multicolumn{2}{c}{Ant-v3} & \multicolumn{2}{c}{HalfCheetah-v3} & \multicolumn{2}{c}{Hopper-v3} & \multicolumn{2}{c}{Walker2d-v3}  \\
 \cmidrule(lr){2-3} \cmidrule(lr){4-5}  \cmidrule(lr){6-7} \cmidrule(lr){8-9} 
 Curriculum (\textbf{\% Online}) & M1 & M2 & M1 & M2 & M1 & M2 & M1 & M2 \\
 \midrule
 \rowfont{\footnotesize} 
 ${C_{\rm add}(t ; f=1M)}$; (\textbf{2.5\%})      & 3093.1 &  894.4 & \textbf{8581.7} & 5275.3 & \textbf{3354.2} & 1977.9 & \textbf{3298.1} & 1882.6 \\
 \rowfont{\scriptsize} 
                                                  & $\pm$ 275.1 &  $\pm$ 113.7 & \textbf{$\pm$ 133.1} & $\pm$ 123.2 & \textbf{$\pm$ 68.3} & $\pm$ 39.6 & \textbf{$\pm$ 144.8} & $\pm$ 56.9 \\
 \rowfont{\footnotesize} 
 ${C_{\rm scale}(t ; c=1.00)}$; (\textbf{2.5\%})  & \textbf{4004.7} & \textbf{2857.0} & \textbf{8417.9} & \textbf{7212.2} & 2712.9 & \textbf{2238.1} & \textbf{3144.9} & \textbf{2050.3} \\
 \rowfont{\scriptsize} 
                                                  & \textbf{$\pm$ 134.8} & \textbf{$\pm$ 150.7} & \textbf{$\pm$ 135.9} & \textbf{$\pm$ 29.2} & $\pm$ 351.5 & \textbf{$\pm$ 110.8} & \textbf{$\pm$ 174.8} & \textbf{$\pm$ 56.6} \\
 \midrule
 \rowfont{\footnotesize} 
 ${C_{\rm add}(t ; f=1M)}$; (\textbf{5.0\%})      & \textbf{3693.9} & 1556.6 & \textbf{8658.8} & 5634.6 & \textbf{3232.1} & 2230.5 & \textbf{3243.8} & \textbf{2097.3}  \\
 \rowfont{\scriptsize} 
                                                  & \textbf{$\pm$ 318.2} & $\pm$ 82.4  & \textbf{$\pm$ 57.9} & $\pm$ 152.7 & \textbf{$\pm$ 180.7} & $\pm$ 71.1 & \textbf{$\pm$ 198.4} & \textbf{$\pm$ 73.0}  \\
 \rowfont{\footnotesize} 
 ${C_{\rm scale}(t ; c=1.00)}$; (\textbf{5.0\%})  & \textbf{3251.1} & \textbf{3068.2} & \textbf{8601.4} & \textbf{7274.0} & \textbf{3356.2}  & \textbf{2459.0} & \textbf{3155.8} & \textbf{2097.0} \\
 \rowfont{\scriptsize} 
                                                  & \textbf{$\pm$ 930.1} & \textbf{$\pm$ 164.2} & \textbf{$\pm$ 53.0} & \textbf{$\pm$ 15.4} & \textbf{$\pm$ 43.8}  & \textbf{$\pm$ 39.1} & \textbf{$\pm$ 184.5} & \textbf{$\pm$ 104.6} \\
 \midrule
 \rowfont{\footnotesize} 
 ${C_{\rm add}(t ; f=1M)}$; (\textbf{10.0\%})     & 4229.0 & 2007.6 & \textbf{8864.2} & 6024.5 & \textbf{2741.2} & 1979.4 & \textbf{3607.7} & \textbf{2448.2} \\
 \rowfont{\scriptsize} 
                                                  & $\pm$ 102.0 & $\pm$ 107.7 & \textbf{$\pm$ 144.3} & $\pm$ 179.7 & \textbf{$\pm$ 613.2} & $\pm$ 447.2 & \textbf{$\pm$ 84.9} & \textbf{$\pm$ 32.2} \\
 \rowfont{\footnotesize} 
 ${C_{\rm scale}(t ; c=1.00)}$; (\textbf{10.0\%}) & \textbf{4510.9} & \textbf{3349.0} & 8608.5 & \textbf{7290.5} & \textbf{3182.9} & \textbf{2580.6} & 3146.9 & 2204.2 \\
 \rowfont{\scriptsize} 
                                                  & \textbf{$\pm$ 158.8} & \textbf{$\pm$ 110.2} & $\pm$ 95.3 & \textbf{$\pm$ 13.6} & \textbf{$\pm$ 218.6} & \textbf{$\pm$ 11.8} & $\pm$ 134.6 & $\pm$ 75.2 \\
 \midrule
 \rowfont{\footnotesize} 
 Teacher Performance & 4876.2 & 3975.8 & 8573.6 & 7285.5 & 3635.2 & 2791.9 & 3927.9 & 2579.8 \\
 \bottomrule \vspace{-0.5em} \\
 \end{tabular}
 \label{tab:results-online-stderr-1}
 \vspace{-1.0em}
\end{table*}

\begin{figure*}[h]
\center
\includegraphics[width=0.85\textwidth]{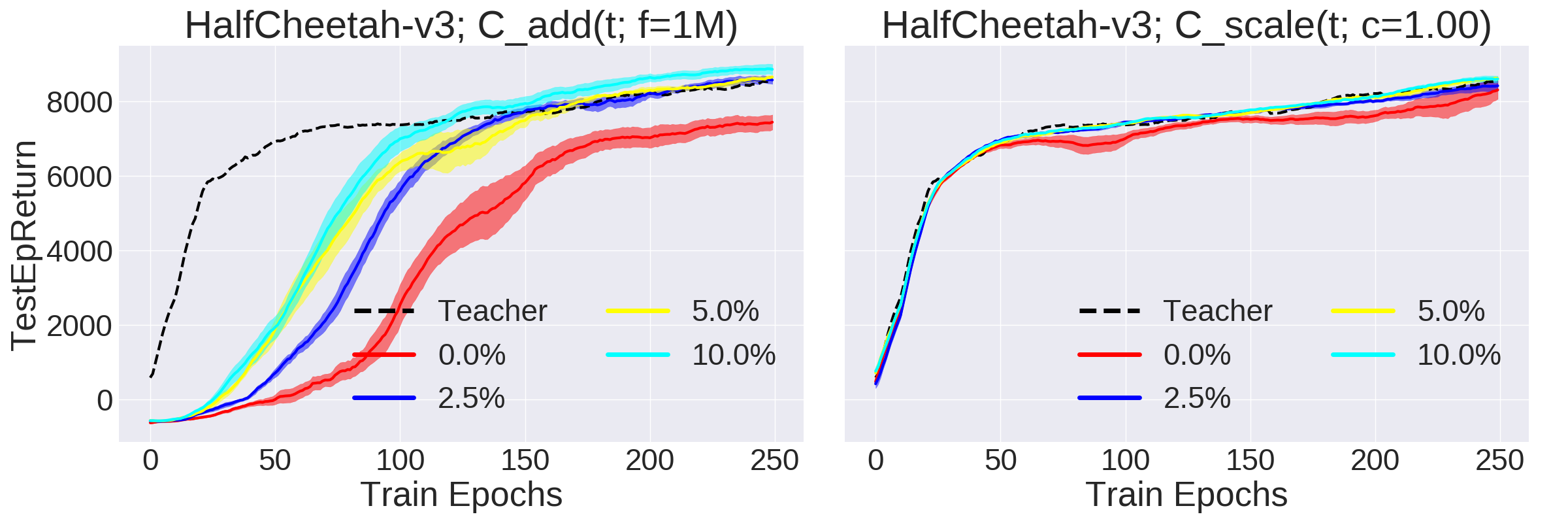}
\caption{
HalfCheetah-v3 test-time performance for students with various amounts of online data, presented in a similar manner as Figure~\ref{fig:online-ant-comparisons}.
}
\vspace*{-10pt}
\label{fig:online-hc-comparisons}
\end{figure*}

\begin{figure*}[h]
\center
\includegraphics[width=0.85\textwidth]{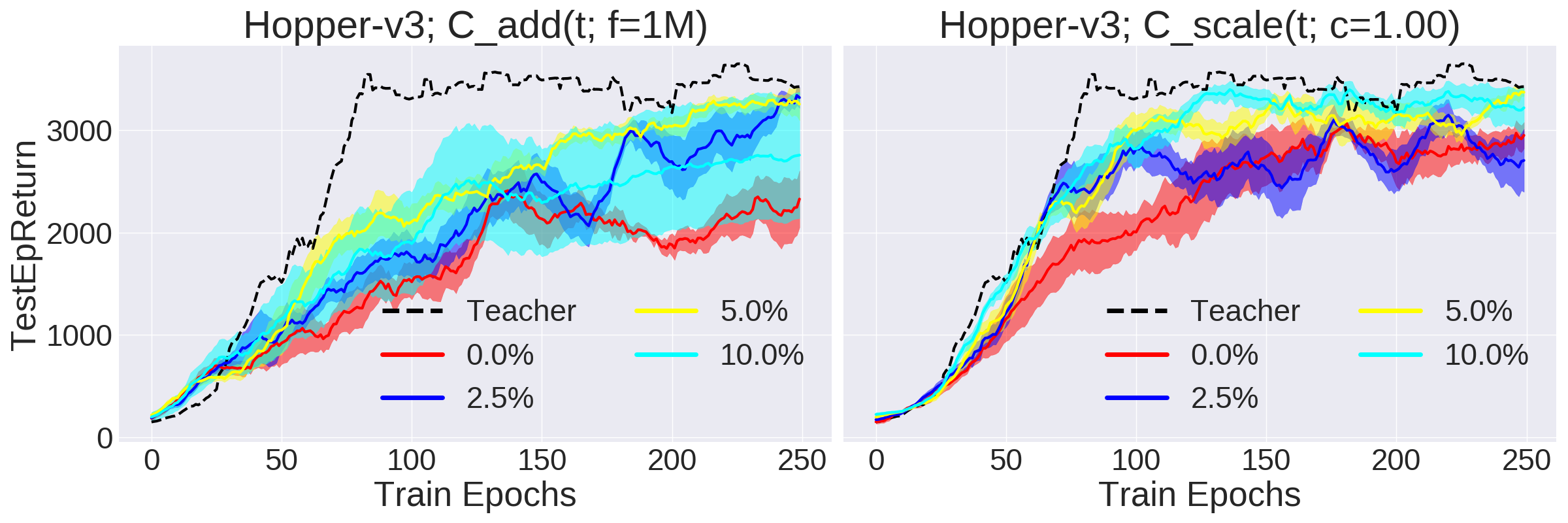}
\caption{
Hopper-v3 test-time performance for students with various amounts of online data, presented in a similar manner as Figure~\ref{fig:online-ant-comparisons}.
}
\vspace*{-10pt}
\label{fig:online-hopper-comparisons}
\end{figure*}

\begin{figure*}[h]
\center
\includegraphics[width=0.85\textwidth]{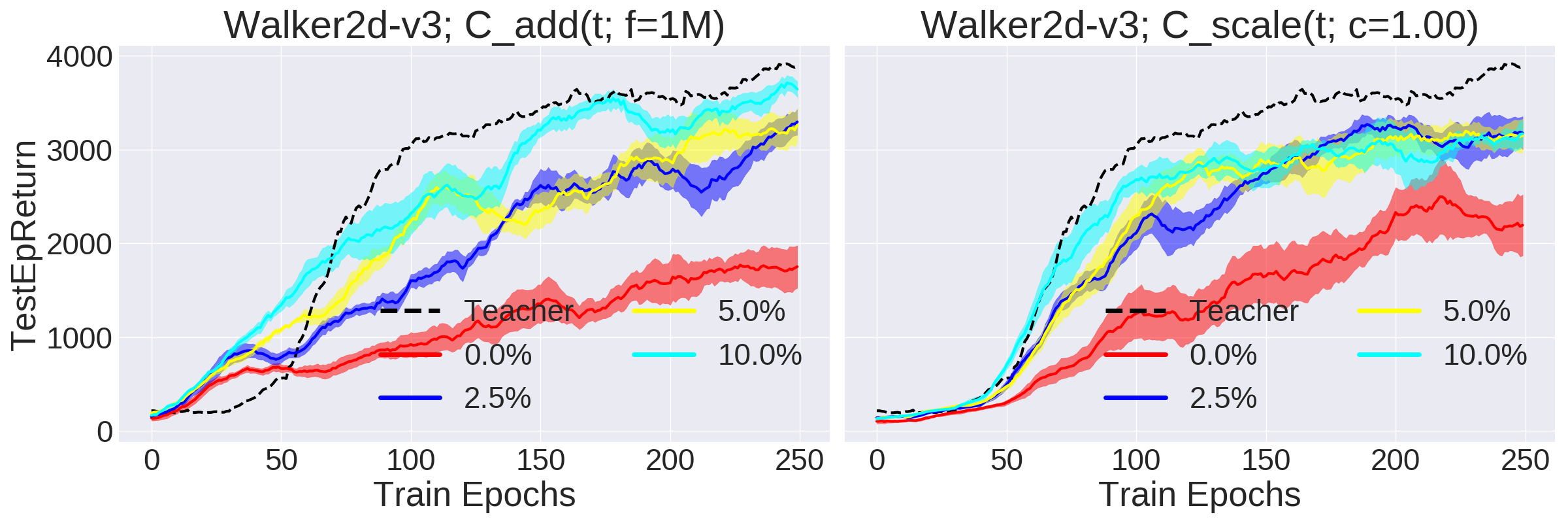}
\caption{
Walker2d-v3 test-time performance for students with various amounts of online data, presented in a similar manner as Figure~\ref{fig:online-ant-comparisons}.
}
\vspace*{-10pt}
\label{fig:online-walker2d-comparisons}
\end{figure*}

In Section~\ref{ssec:online-learning}, we report results from adding a small amount of online data to the TD3 students, to better mitigate the difficulty of offline learning (particularly with Ant-v3) but also to understand if the curriculum continues to help.
Table~\ref{tab:results-online-stderr-1} lists the results. As expected, an increase in the number of self-generated samples from 2.5\% to 10.0\% results in an increase in student performance. We also see that despite only getting 2.5\% of online samples, off-the-shelf TD3 students can avoid collapse in Ant-v3, and that performance further increases from utilizing the $C_{\rm scale}(t; c=1.00)$ curriculum over $C_{\rm add}(t; f=1M)$. 
Figures~\ref{fig:online-ant-comparisons},~\ref{fig:online-hc-comparisons},~\ref{fig:online-hopper-comparisons}, and~\ref{fig:online-walker2d-comparisons} show the plots for the respective environments: Ant-v3, HalfCheetah-v3, Hopper-v3, and Walker2d-v3.

\subsection{More Detailed Results from Section~\ref{ssec:q-values} and Other Diagnostics Investigation}\label{app:q-values}

\begin{figure*}[t]
\center
\includegraphics[width=1.00\textwidth]{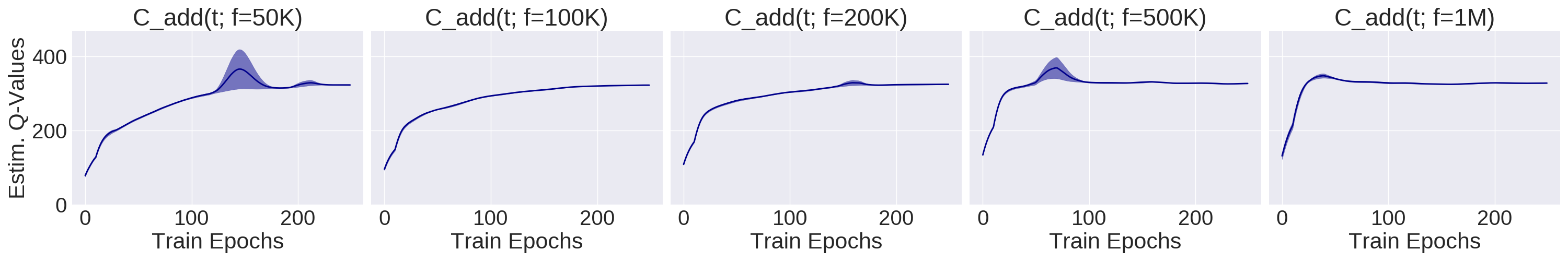}
\caption{
Students' estimated Q-values in Hopper-v3, presented in the similar manner as Figure~\ref{fig:results-qvalues-hc}.
}
\vspace*{-5pt}
\label{fig:results-qvalues-hopper}
\end{figure*}

\begin{figure*}[t]
\center
\includegraphics[width=1.00\textwidth]{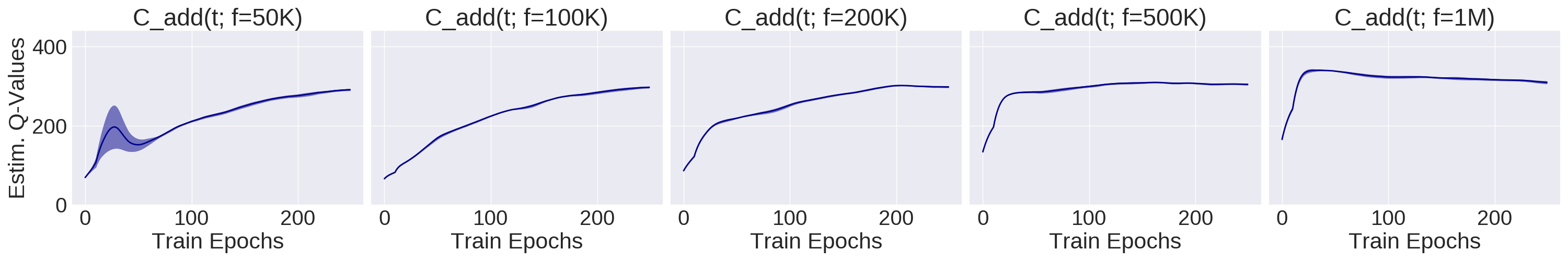}
\caption{
Students' estimated Q-values in Walker2d-v3, presented in the same manner as Figure~\ref{fig:results-qvalues-hc}.
}
\vspace*{-5pt}
\label{fig:results-qvalues-walker2d}
\end{figure*}

\begin{figure*}[t]
\center
\includegraphics[width=1.00\textwidth]{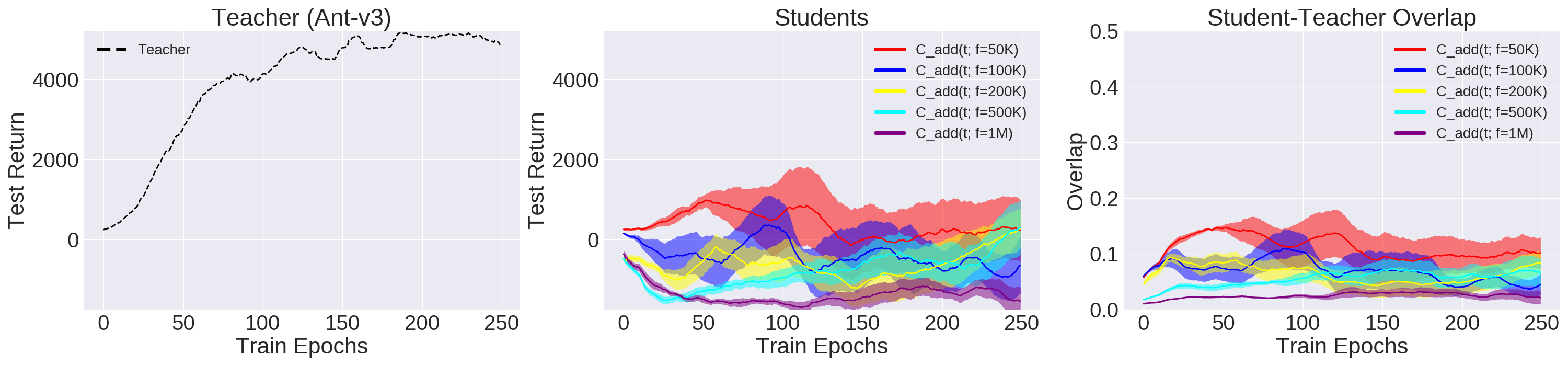}
\caption{
Plots for Ant-v3 showing teacher test-time performance (left), student performance curricula (middle), and the overlap between these students and the teacher dataset (right, same color code).
}
\vspace*{-5pt}
\label{fig:overlap-ant-logged}
\end{figure*}

\begin{figure*}[t]
\center
\includegraphics[width=1.00\textwidth]{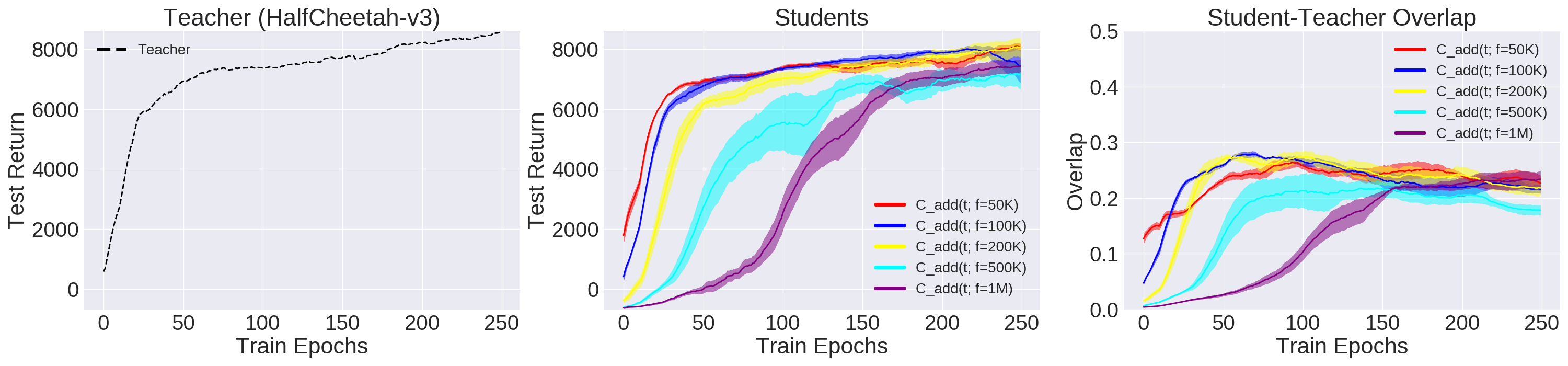}
\caption{
Plots for HalfCheetah-v3 showing teacher test-time performance (left), student performance curricula (middle), and the overlap between these students and the teacher dataset (right, same color code).
}
\vspace*{-5pt}
\label{fig:overlap-hc-logged}
\end{figure*}

\begin{figure*}[t]
\center
\includegraphics[width=1.00\textwidth]{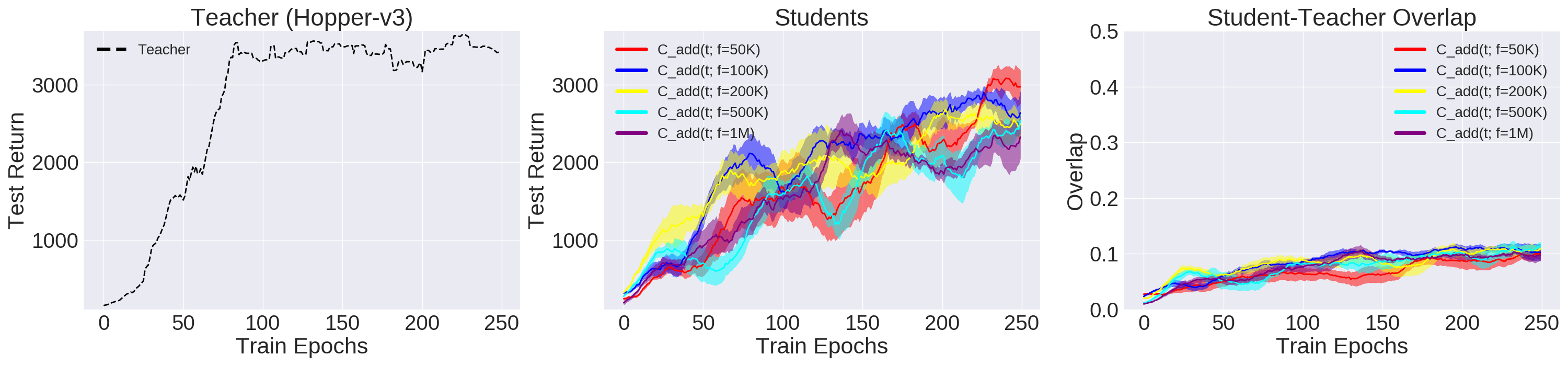}
\caption{
Plots for Hopper-v3 showing teacher test-time performance (left), student performance curricula (middle), and the overlap between these students and the teacher dataset (right, same color code).
}
\vspace*{-5pt}
\label{fig:overlap-hopper-logged}
\end{figure*}

\begin{figure*}[t]
\center
\includegraphics[width=1.00\textwidth]{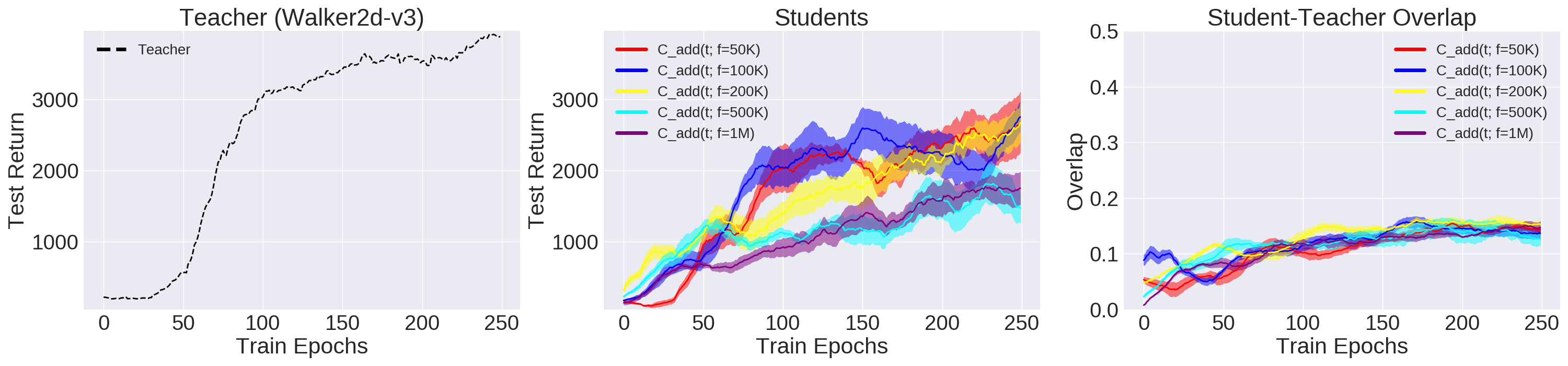}
\caption{
Plots for Walker2d-v3 showing teacher test-time performance (left), student performance curricula (middle), and the overlap between these students and the teacher dataset (right, same color code).
}
\vspace*{-5pt}
\label{fig:overlap-walker2d-logged}
\end{figure*}

Figures~\ref{fig:results-qvalues-hc},~\ref{fig:results-qvalues-hopper}, and~\ref{fig:results-qvalues-walker2d} show plots of estimated Q-values from various additive curricula for HalfCheetah-v3, Hopper-v3, and Walker2d-v3, respectively. These are from the offline TD3 students with 250 epochs of training (Section~\ref{ssec:results-250-epochs}). We do not present the Ant-v3 results as the estimated Q-values diverged. The results for HalfCheetah-v3, Hopper-v3, and Walker2d-v3 indicate that estimated Q-values tend to have a notable early upwards ``bump'' with too much data available for initial learning, which can hinder training.

For these same students, Figures~\ref{fig:overlap-ant-logged},~\ref{fig:overlap-hc-logged},~\ref{fig:overlap-hopper-logged}, and~\ref{fig:overlap-walker2d-logged} show overlap plots for additive curricula for Ant-v3, HalfCheetah-v3, Hopper-v3, and Walker2d-v3, respectively. We observe that overlap is generally a good predictor of student performance and that overlap values mirror the learning curves. In future work, we will investigate an application of overlap for selecting among a set of distinct teacher datasets, and see if picking the one with highest overlap with the current student policy leads to faster student learning.

\subsection{Soft Actor-Critic Students and Teachers}\label{ssec:sac-results}

\begin{table*}[t]
 \setlength\tabcolsep{5.0pt}
 \caption{
 \textbf{\alg Results for SAC.} We use the logged environment history of SAC teachers to create the data for each environment. Then, we use SAC again for students, and test three data curricula. Standard errors are reported in separate rows (after the $\pm$ symbols). For more details, see Section~\ref{ssec:sac-results}.
 }
 \centering
 \footnotesize
 \begin{tabular}{@{}LRRRRRRRR@{}} 
 \toprule
 & \multicolumn{2}{c}{Ant-v3} & \multicolumn{2}{c}{HalfCheetah-v3} & \multicolumn{2}{c}{Hopper-v3} & \multicolumn{2}{c}{Walker2d-v3}  \\
 \cmidrule(lr){2-3} \cmidrule(lr){4-5} \cmidrule(lr){6-7} \cmidrule(lr){8-9}
 Curriculum & M1 & M2 & M1 & M2 & M1 & M2 & M1 & M2 \\
 \midrule
 \rowfont{\footnotesize}
 ${C_{\rm add}(t ; f=1M)}$ &  -2929.1 & -2509.5 &   706.4 & 340.3  &  2012.5 & 1567.4 &  2559.6 & 1534.0 \\
 \rowfont{\scriptsize}
                           & $\pm$ 51.4 & $\pm$ 38.3 &   $\pm$ 131.8  & $\pm$ 35.3 &  $\pm$ 304.4 & $\pm$ 171.1 &  $\pm$ 362.8  & $\pm$ 164.4 \\
 \rowfont{\footnotesize}
 ${C_{\rm scale}(t ; c=1.00)}$  &  \textbf{2202.9} & \textbf{2615.4} & \textbf{11913.5} & 10017.9 &  \textbf{3598.5} & \textbf{2258.8} &  3090.0 & 2142.3 \\
 \rowfont{\scriptsize}
                                &  \textbf{$\pm$ 1529.8} & \textbf{$\pm$ 179.9} & \textbf{$\pm$ 62.0} & $\pm$ 53.7 &  \textbf{$\pm$ 16.4} & \textbf{$\pm$ 126.7} &  $\pm$ 647.7  & $\pm$ 113.7 \\
 \rowfont{\footnotesize}
 ${C_{\rm scale}(t ; c=1.25)}$ &  \textbf{177.7} & \textbf{2533.8} & 11685.3 & \textbf{10120.4} &  3196.2 & \textbf{2359.7} & \textbf{4417.3} & \textbf{2631.4} \\
 \rowfont{\scriptsize}
                               &  \textbf{$\pm$ 1698.8} & \textbf{$\pm$ 261.5} & $\pm$ 74.3 & \textbf{$\pm$ 44.5} &  $\pm$ 137.2 & \textbf{$\pm$ 81.1} & \textbf{$\pm$ 86.1} & \textbf{$\pm$ 46.1} \\
 \midrule
 \rowfont{\footnotesize}
  SAC Teacher & 5556.5 & 3831.5 & 11847.7 & 10258.8 & 3665.3 & 3002.2 & 4408.1 & 2651.2 \\
 \bottomrule \vspace{-0.5em} \\
 \end{tabular}
 \label{tab:results-sac}
 \vspace{-0.5em}
\end{table*}

We next perform experiments using SAC~\cite{sac} for teachers and students, to see if results from the data curriculum generalize across Deep RL algorithms. We train SAC to build new teacher datasets for each of the four MuJoCo environments in a similar manner as we did for TD3. Then, we train SAC students from the teacher data with the same 5 student random seeds (90 through 94), using 3 data curricula.\footnote{We test fewer data curricula as compared to TD3-to-TD3 teaching due to compute limits.}
The results, in Table~\ref{tab:results-sac}, indeed suggest that data curriculum matters for SAC-to-SAC teaching. The baseline curriculum of using all the data, $C_{\rm add}(t; f=1M)$, results in the worst student performance by far among the 3 tested curricula, with no top performances in any of the environments by either the M1 or M2 metrics. In contrast, the two other data curricula, $C_{\rm scale}(t; c=1.00)$ and $C_{\rm scale}(t; c=1.25)$, limit the available range of samples and have comparable performance.

\subsection{Cross-Algorithm Teaching}\label{app:cross-alg-teaching}

In experiments, we have utilized TD3 teachers and students, and have also tested SAC teacher and students (Section~\ref{ssec:sac-results}).
We next try \emph{cross-algorithm teaching}: going from TD3 teachers to SAC students, and vice versa. This is to test whether it is still possible to learn offline using a different algorithm than the one which generated the data, and whether a data curriculum also helps accelerate learning.
To facilitate comparisons with prior experiments, we keep settings as consistent as possible, with the same 8 teacher datasets across the 4 environments, and the same random seeds for the new students (90 through 94). 
Tables~\ref{tab:results-cross-alg-td3-to-sac} and~\ref{tab:results-cross-alg-sac-to-td3} show the results for TD3-to-SAC teaching and SAC-to-TD3 teaching, respectively. We bold face results using the same convention of overlapping standard errors as described in Section~\ref{app:standard_error_overlap}.

The results suggest that the choice of data curriculum can continue to have major effects on the performance of the student, particularly with the SAC-to-TD3 teaching setting, where curricula that allow too much data result in worse student performance. Thus, a proper selection of the data curriculum can improve learning across different algorithms, increasing the generality of the \alg framework.
We caution that results for TD3-to-SAC seem more mixed, particularly for Ant-v3 and Hopper-v3, which exhibit extraordinarily high variance. Furthermore, Hopper-v3 with SAC students and TD3 teachers appears to benefit from having more data, instead of less data, throughout training.

We also run experiments using TD3 for teachers and train students using Batch Constrained Q-learning (BCQ)~\cite{BCQ_2019}, to see if the curriculum helps with learning for an algorithm specialized to offline reinforcement learning. Due to compute limitations, we use 2 random seeds (instead of 5) for each student, and we test 2 curricula (one of which is the baseline of using all the data at all times).
Table~\ref{tab:results-td3-to-bcq} contains the results. Overall, unlike prior experiments with standard off-policy algorithms, the results suggest that the choice of curriculum has less of an effect on BCQ. The baseline curriculum of always allowing all data performs as well as $C_{\rm add}(t; f=50K)$ for HalfCheetah-v3 and Hopper-v3 according to both M1 and M2. In future work we will explore this in more detail.

\begin{table*}[t]
 \setlength\tabcolsep{5.0pt}
 \caption{
 \textbf{\alg, TD3 to SAC Cross-Algorithm Teaching}. We report experiments where TD3 teachers generate the data (results in the bottom row as reference) and where SAC students with different data curricula learn offline from the resulting data. Standard errors of the average reward (among the 5 seeds) are reported in smaller font. See Section~\ref{app:cross-alg-teaching} for further details.
 }
 \centering
 \footnotesize
 \begin{tabular}{@{}LRRRRRRRR@{}} 
 \toprule
 & \multicolumn{2}{c}{Ant-v3} & \multicolumn{2}{c}{HalfCheetah-v3} & \multicolumn{2}{c}{Hopper-v3} & \multicolumn{2}{c}{Walker2d-v3}  \\
 \cmidrule(lr){2-3} \cmidrule(lr){4-5} \cmidrule(lr){6-7} \cmidrule(lr){8-9}
 Data Curriculum & M1 & M2 & M1 & M2 & M1 & M2 & M1 & M2 \\
 \midrule
 \rowfont{\footnotesize}
 ${C_{\rm add}(t ; f=50K)}$    & \textbf{-1737.1} & -145.0  &  8679.9 & 7598.2 & 5.2 & 82.0 & \textbf{3481.4} & 1942.9 \\
 \rowfont{\scriptsize}
                               & \textbf{$\pm$ 803.5} & $\pm$ 500.1  & $\pm$ 180.3 & $\pm$ 21.9 & $\pm$ 2.6 & $\pm$ 2.9 & \textbf{$\pm$ 220.9} & $\pm$ 357.6 \\
 \rowfont{\footnotesize}
 ${C_{\rm add}(t ; f=100K)}$   & \textbf{-2240.5} & -1739.0 & 8785.4 & 7469.2 &  \textbf{2185.0} & \textbf{1493.1} & \textbf{3615.3} & \textbf{2897.1} \\
 \rowfont{\scriptsize}
                               & \textbf{$\pm$ 469.8} & $\pm$ 228.0  &  $\pm$ 38.6 & $\pm$ 18.0  & \textbf{$\pm$ 718.5} & \textbf{$\pm$ 493.7} & \textbf{$\pm$ 150.2} & \textbf{$\pm$ 42.3} \\
 \rowfont{\footnotesize}
 ${C_{\rm add}(t ; f=200K)}$   & -2977.4  & -1985.3  & 8603.7  & 7022.4 & 1314.5 & \textbf{1257.7} & \textbf{3431.5} & \textbf{2885.8} \\
 \rowfont{\scriptsize}
                               & $\pm$ 13.1 & $\pm$ 52.7  & $\pm$ 70.6 & $\pm$ 104.4  & $\pm$ 494.2 & \textbf{$\pm$ 406.1} & \textbf{$\pm$ 148.3} & \textbf{$\pm$ 91.6}  \\
 \rowfont{\footnotesize}
 ${C_{\rm add}(t ; f=500K)}$   & \textbf{-1505.8} & -1332.3  & 8275.1 & 5518.9 & \textbf{2553.2} & \textbf{1496.2} & 3207.4 & 2170.7 \\
 \rowfont{\scriptsize}
                               & \textbf{$\pm$ 1220.5} & $\pm$ 543.4 & $\pm$ 98.6 & $\pm$ 97.3 & \textbf{$\pm$ 213.9} & \textbf{$\pm$ 151.0} & $\pm$ 241.6 & $\pm$ 145.6 \\
 \rowfont{\footnotesize}
 ${C_{\rm add}(t ; f=1M)}$     & -2891.2  & -1742.0  &  8301.1  & 4282.4  & \textbf{2053.6} & \textbf{1325.2} & 3241.2 & 2316.5 \\
 \rowfont{\scriptsize}
                               & $\pm$ 66.7 & $\pm$ 128.6  &  $\pm$ 244.4 & $\pm$ 217.1 & \textbf{$\pm$ 555.3} & \textbf{$\pm$ 316.9} & $\pm$ 120.9 & $\pm$ 65.8 \\
 \midrule
 \rowfont{\footnotesize}
 ${C_{\rm scale}(t ; c=0.50)}$ & \textbf{-753.8} & -412.6  & 7420.0 & 6455.2 &  76.6 & 189.1 & 1931.1 & 760.7  \\
 \rowfont{\scriptsize}
                               & \textbf{$\pm$ 1153.7} & $\pm$ 447.9  &  $\pm$ 118.3 & $\pm$ 58.9 & $\pm$ 44.9 & $\pm$ 31.3  & $\pm$ 598.9 & $\pm$ 178.2 \\
 \rowfont{\footnotesize}
 ${C_{\rm scale}(t ; c=0.75)}$ & \textbf{-537.2} & \textbf{857.2}  & 8423.7 & 6928.9 &  6.0 & 148.3 & 2362.7 & 1397.7 \\
 \rowfont{\scriptsize}
                               & \textbf{$\pm$ 1389.3} & \textbf{$\pm$ 630.9}  & $\pm$ 47.3 & $\pm$ 45.8 & $\pm$ 1.4 & $\pm$ 7.6 & $\pm$ 649.1 & $\pm$ 295.7 \\
 \rowfont{\footnotesize}
 ${C_{\rm scale}(t ; c=1.00)}$ & \textbf{-1258.0} & \textbf{1318.5}  & \textbf{9192.2} & 7461.6 & 57.9 & 117.6 & \textbf{3344.5} & 2322.4 \\
 \rowfont{\scriptsize}
                               & \textbf{$\pm$ 1356.5} & \textbf{$\pm$ 602.5}  & \textbf{$\pm$ 50.5} & $\pm$ 66.0 & $\pm$ 32.7 & $\pm$ 4.4  & \textbf{$\pm$ 392.3} & $\pm$ 216.6 \\
 \rowfont{\footnotesize}
 ${C_{\rm scale}(t ; c=1.10)}$ & \textbf{-1339.7} & \textbf{1137.2}  & 8666.9 & 7440.0  & 5.9 & 94.3 & \textbf{3215.4} & 2369.1 \\
 \rowfont{\scriptsize}
                               & \textbf{$\pm$ 1387.6} & \textbf{$\pm$ 534.2}  & $\pm$ 391.4 & $\pm$ 96.7 & $\pm$ 1.6 & $\pm$ 6.6 & \textbf{$\pm$ 288.8} & $\pm$ 95.9 \\
 \rowfont{\footnotesize}
 ${C_{\rm scale}(t ; c=1.25)}$ & \textbf{-1036.7} & \textbf{1215.6}  & \textbf{9182.1} & \textbf{7731.5}  & 14.2 & 105.1 & \textbf{3366.6} & 2686.7 \\
 \rowfont{\scriptsize}
                               & \textbf{$\pm$ 1262.7} & \textbf{$\pm$ 559.0}  & \textbf{$\pm$ 54.2} & \textbf{$\pm$ 31.4} & $\pm$ 9.5 & $\pm$ 10.4 & \textbf{$\pm$ 294.4} & $\pm$ 51.3 \\
 \midrule
 \rowfont{\footnotesize}
 TD3 Teacher     & 4876.2 & 3975.8 & 8573.6 & 7285.5 & 3635.2 & 2791.9 & 3927.9 & 2579.8 \\
 \bottomrule \vspace{-0.5em} \\
 \end{tabular}
 \label{tab:results-cross-alg-td3-to-sac}
 \vspace{-0.5em}
\end{table*}

\begin{table*}[t]
 \setlength\tabcolsep{5.0pt}
 \caption{
 \textbf{\alg, SAC to TD3 Cross-Algorithm Teaching}. We report experiments where SAC teachers generate the data (results in the bottom row as reference) and where TD3 students with different data curricula learn offline from the resulting data. Standard errors of the average reward (among the 5 seeds) are reported in smaller font. See Section~\ref{app:cross-alg-teaching} for further details.
 }
 \centering
 \footnotesize
 \begin{tabular}{@{}LRRRRRRRR@{}} 
 \toprule
 & \multicolumn{2}{c}{Ant-v3} & \multicolumn{2}{c}{HalfCheetah-v3} & \multicolumn{2}{c}{Hopper-v3} & \multicolumn{2}{c}{Walker2d-v3}  \\
 \cmidrule(lr){2-3} \cmidrule(lr){4-5} \cmidrule(lr){6-7} \cmidrule(lr){8-9}
 Data Curriculum & M1 & M2 & M1 & M2 & M1 & M2 & M1 & M2 \\
 \midrule
 \rowfont{\footnotesize}
 ${C_{\rm add}(t ; f=50K)}$     & \textbf{183.6} & \textbf{-208.3}  &  8021.3 & 7529.8 & \textbf{1708.6} & \textbf{1333.6} & \textbf{3024.1} & \textbf{1557.2} \\
 \rowfont{\scriptsize}
                                & \textbf{$\pm$ 734.7} & \textbf{$\pm$ 391.0}  & \textbf{$\pm$ 537.1} & $\pm$ 201.9 & \textbf{$\pm$ 324.1} & \textbf{$\pm$ 122.1} & \textbf{$\pm$ 261.8} & \textbf{$\pm$ 190.4} \\
 \rowfont{\footnotesize}
 ${C_{\rm add}(t ; f=100K)}$    & -2184.5 &  -447.6 &  4544.6 & 5429.8  & \textbf{1975.4} & \textbf{1363.9} & \textbf{2955.1} & 1630.6 \\
 \rowfont{\scriptsize}
                                & $\pm$ 291.2 & $\pm$ 439.8 &  $\pm$ 716.2 & $\pm$ 249.5 & \textbf{$\pm$ 318.3} & \textbf{$\pm$ 166.5} & \textbf{$\pm$ 176.2} & $\pm$ 70.6 \\
 \rowfont{\footnotesize}
 ${C_{\rm add}(t ; f=200K)}$    & -1345.4 & -1405.8 & 2966.7 & 2417.6 & 1518.5 & \textbf{1136.2} & \textbf{2685.7} & \textbf{1742.1} \\
 \rowfont{\scriptsize}
                                & $\pm$ 649.0 & $\pm$ 333.7 & $\pm$ 731.8 & $\pm$ 256.0 & $\pm$ 304.7 & \textbf{$\pm$ 92.1} & \textbf{$\pm$ 418.5} & \textbf{$\pm$ 92.4} \\
 \rowfont{\footnotesize}
 ${C_{\rm add}(t ; f=500K)}$    & -2182.3 & -2324.2 & 493.3 & 265.9 & 1327.6 & 913.3  & 2361.2 & 1615.0 \\
 \rowfont{\scriptsize}
                                & $\pm$ 421.0 & $\pm$ 163.0 & $\pm$ 163.7 & $\pm$ 72.0 & $\pm$ 327.9 & $\pm$ 57.3 & $\pm$ 349.8 & $\pm$ 119.9  \\
 \rowfont{\footnotesize}
 ${C_{\rm add}(t ; f=1M)}$      & -1920.9 & -2162.6 & -173.3 & -228.0 & 1612.1 & 1006.7  & 2328.5 & 1179.5  \\
 \rowfont{\scriptsize}
                                & $\pm$ 280.1 & $\pm$ 139.5 & $\pm$ 90.1 & $\pm$ 44.8 & $\pm$ 220.4 & $\pm$ 44.8 & $\pm$ 293.3 & $\pm$ 30.9 \\
 \midrule
 \rowfont{\footnotesize}
 ${C_{\rm scale}(t ; c=0.50)}$  & \textbf{663.7} & \textbf{309.8}  & \textbf{9786.1} & 7434.4 &  \textbf{2130.8} & \textbf{1185.7} &  1049.7 & 474.2 \\
 \rowfont{\scriptsize}
                                & \textbf{$\pm$ 412.6} & \textbf{$\pm$ 147.6}  & \textbf{$\pm$ 150.2} & $\pm$ 48.8  &  \textbf{$\pm$ 271.9} & \textbf{$\pm$ 90.5} & $\pm$ 265.8 & $\pm$ 79.9  \\
 \rowfont{\footnotesize}
 ${C_{\rm scale}(t ; c=0.75)}$ & -2016.4 & -280.4 & 9130.3 & \textbf{7729.7} & 1454.5  & \textbf{1257.3} & 2544.9 & 1076.3 \\
 \rowfont{\scriptsize}
                               & $\pm$ 388.6 & $\pm$ 354.4 & $\pm$ 301.3 & \textbf{$\pm$ 139.0} & $\pm$ 188.5 & \textbf{$\pm$ 154.2} & $\pm$ 120.4 & $\pm$ 50.2 \\
 \rowfont{\footnotesize}
 ${C_{\rm scale}(t ; c=1.00)}$ & -1851.5 & \textbf{-153.8} & 7179.5 & 7473.5 & 1164.4  & \textbf{1230.1} & \textbf{2551.6} & 1329.6 \\
 \rowfont{\scriptsize}
                               & $\pm$ 747.7 & \textbf{$\pm$ 319.2} & $\pm$ 595.4 & $\pm$ 108.7 & $\pm$ 292.4 & \textbf{$\pm$ 85.6} & \textbf{$\pm$ 322.0} & $\pm$ 56.6 \\
 \rowfont{\footnotesize}
 ${C_{\rm scale}(t ; c=1.10)}$ & -1501.1 & -408.2 & 8405.4 & \textbf{7814.7} & \textbf{1718.6} & \textbf{1150.9} & \textbf{2847.5} & 1648.2  \\
 \rowfont{\scriptsize}
                               & $\pm$ 1107.9 & $\pm$ 551.6 & $\pm$ 317.4 & \textbf{$\pm$ 35.8} & \textbf{$\pm$ 342.3} & \textbf{$\pm$ 192.8}  & \textbf{$\pm$ 183.5} & $\pm$ 48.1 \\
 \rowfont{\footnotesize}
 ${C_{\rm scale}(t ; c=1.25)}$ & -1274.4 & \textbf{77.6}  & 7662.7 & 7499.9 & \textbf{1802.2} & \textbf{1238.4} &  \textbf{2896.6} & \textbf{1790.0} \\
 \rowfont{\scriptsize}
                               & $\pm$ 757.1 & \textbf{$\pm$ 543.6}  & $\pm$ 207.4 & $\pm$ 99.5  & \textbf{$\pm$ 372.6} & \textbf{$\pm$ 103.5} & \textbf{$\pm$ 168.9} & \textbf{$\pm$ 45.3} \\
 \midrule
 \rowfont{\footnotesize}
 SAC Teacher & 5556.5 & 3831.5 & 11847.7 & 10258.8 & 3665.3 & 3002.2 & 4408.1 & 2651.2 \\
 \bottomrule \vspace{-0.5em} \\
 \end{tabular}
 \label{tab:results-cross-alg-sac-to-td3}
 \vspace{-0.5em}
\end{table*}

\begin{table*}[t]
 \setlength\tabcolsep{5.0pt}
 \caption{
 \textbf{\alg, TD3 to BCQ Cross-Algorithm Teaching.} We train students using BCQ, with TD3 teachers, and test two data curricula. Standard errors are reported in separate rows (after the $\pm$ symbols). See Section~\ref{app:cross-alg-teaching} for further details.
 }
 \centering
 \footnotesize
 \begin{tabular}{@{}LRRRRRRRR@{}} 
 \toprule
 & \multicolumn{2}{c}{Ant-v3} & \multicolumn{2}{c}{HalfCheetah-v3} & \multicolumn{2}{c}{Hopper-v3} & \multicolumn{2}{c}{Walker2d-v3}  \\
 \cmidrule(lr){2-3} \cmidrule(lr){4-5} \cmidrule(lr){6-7} \cmidrule(lr){8-9}
 Curriculum & M1 & M2 & M1 & M2 & M1 & M2 & M1 & M2 \\
 \midrule
 \rowfont{\footnotesize}
 ${C_{\rm add}(t ; f=50K)}$ &  \textbf{5245.6} & \textbf{5022.8} & \textbf{5365.0} & \textbf{5512.2} &  \textbf{477.8} & \textbf{508.8} & \textbf{1711.3} & \textbf{875.7} \\
 \rowfont{\scriptsize}
                           & \textbf{$\pm$ 10.8} & \textbf{$\pm$ 17.5} &   \textbf{$\pm$ 558.5}  & \textbf{$\pm$ 378.3} &  \textbf{$\pm$ 105.4} & \textbf{$\pm$ 94.7} & \textbf{$\pm$ 467.1} & \textbf{$\pm$ 219.3} \\
 \rowfont{\footnotesize}
 ${C_{\rm add}(t ; f=1M)}$  &  5137.9 & \textbf{5019.6} & \textbf{5158.6} & \textbf{5292.8} &  \textbf{721.3} & \textbf{403.5} &  609.5 & 549.7 \\
 \rowfont{\scriptsize}
                                &  $\pm$ 42.9 & \textbf{$\pm$ 52.1} & \textbf{$\pm$ 75.8} & \textbf{$\pm$ 150.3} &  \textbf{$\pm$ 513.2} & \textbf{$\pm$ 75.9} &  $\pm$ 235.9  & $\pm$ 48.7 \\
 \midrule
 \rowfont{\footnotesize}
  TD3 Teacher & 4876.2 & 3929.0  & 8573.6 &  7191.7 & 3635.2 & 2759.1 & 3813.3 & 2321.8\\
 \bottomrule \vspace{-0.5em} \\
 \end{tabular}
 \label{tab:results-td3-to-bcq}
 \vspace{-1.0em}
\end{table*}


\section{Other Things We Tried}\label{app:did-not-work}

This section reports things we have tried which gave inconclusive or mixed results. We report these to spur research interest into why we observe these phenomena and to potentially save research effort and compute for those who may be interested in building upon this study.

\subsection{Time-Based Reward Shaping}\label{ssec:time-reward}

To accelerate student learning, we propose reward shaping by utilizing a time-based predictor ${h_\psi: \mathcal{S} \to [0,1]}$ which is a function mapping a state $\bs \in \mathcal{S}$ to a value between 0 and 1, where the value is interpreted as the level of ``advancement.'' Given the teacher data $\tbuf$, for each time index $i$, we train $h_\psi(\bs_i)$ to predict label $i / |\tbuf|$ via regression with a mean square error loss. In this paper, $|\tbuf| = 1M$, so each data tuple $(\bs_i, \ba_i, r_i, \bs_{i+1})$ is associated with an index $i$ between 0 and 1M (exclusive). Given a time predictor, we can adjust the reward for a given tuple as follows:
\begin{equation}\label{eq:time-reward-shaping}
    (\bs_i, \ba_i, r_i, \bs_{i+1}) \Longrightarrow
    (\bs_i, \ba_i, \underbrace{r_i + \alpha (h_\psi(\bs_{i+1}) - h_\psi(\bs_i))}_{\rm Reward}, \bs_{i+1}),
\end{equation}
where $\alpha > 0$ is a hyperparameter.
Intuitively, this will add an extra reward bonus which encourages the student to prioritize actions that take it from a ``less advanced'' state to a ``more advanced'' state.
While this idea seems appealing, so far we have not seen clear benefits from utilizing the revised data tuples in Equation~\ref{eq:time-reward-shaping}, nor have we see clear benefits if we modify the reward by only adding $\alpha \cdot h_\psi(\bs_{i+1})$ and ignoring the ``baseline'' term $h_\psi(\bs_i)$. One reason might be that the dense, hand-designed MuJoCo environment rewards are already well suited for inducing desired behavior, and hence it would be difficult to design better rewards.

We also attempted a sparse-reward variant to see if students can learn from ignoring the reward from the data tuples by setting all $r_i=0$ in $\tbuf$ and just using time predictor as a signal. This may be a more realistic offline learning scenario in that a precise numerical reward from the teacher data is not necessary. However, we observed poor results, and hence we do not report these experiments.

\subsection{Reordering Tuples Based on Reward}\label{ssec:reorder-tuples}

The data $\tbuf$ comes from the teacher's experienced samples from its training history. Thus there is a natural ordering in $\tbuf$, in which earlier data tuples (\ie smaller time indices) tend to have lower reward on average than later data tuples due to the progression of teacher training. The curricula we test in the paper all rely on restricting the range of time indices in $\tbuf$ which are accessible to the student, and in particular, they result in a range of \emph{consecutive} time indices.  We hypothesized that the stronger performance of curricula that allow the available data in $\tbuf$ to grow over time could be due to the increasing reward (on average) from the data tuples. Thus, we tried first reordering \emph{all} 1M data tuples in $\tbuf$ to form a newer dataset where all data tuples are ordered by increasing reward, in that the following data indexing relation holds:
\begin{equation}
\begin{split}
\forall \; 0 \le i < j <1M, \;\; \mbox{we have}\;\; 
r_i \le r_j \;\; \mbox{in tuples} \;\; \\
(\bs_i, \ba_i, r_i, \bs_{i+1}) \;\; \mbox{and} \;\;
(\bs_j, \ba_j, r_j, \bs_{j+1}).
\end{split}
\end{equation}
We experimented with the two primary curricula from the main part of the paper, $C_{\rm add}(t; f=1M)$ and $C_{\rm scale}(t; c=1.00)$.
However, this reordering of the buffer did not result in performance gains over simply keeping the original data ordering in $\tbuf$, suggesting that there is something more fundamental than just the raw reward. We also tried ranking data tuples according to the \emph{episodic} reward, and then reordering based on that, but this did not yield gains.

\subsection{Roll out Snapshot Plus Noise}\label{ssec:rollout-noise}

Instead of using the logged environment history from a teacher's training run, 
another option is to take the final teacher policy after training, $\pi_{\theta_T}$, and roll it out in the environment with noise added for diverse coverage to generate 1M data tuples for $\tbuf$. We have tested two methods for adding noise to $\pi_{\theta_T}$, which use Gaussian noise and uniform noise.

In the case with Gaussian noise, we define two parameters. First, we use a parameter $\Lambda$ that specifies a maximum possible standard deviation. Second, a parameter $\xi$ represents the probability that we inject noise at any time. Given these two parameters, \emph{for each episode} $k$, we sample a standard deviation ${\sigma_k \sim {\rm Unif}(0, \Lambda)}$ via a uniform distribution. Then, with this $\sigma_k$ \emph{fixed} for all time steps in the episode, the teacher's rollout policy is defined as:
\begin{equation}
    \ba_t = 
    \begin{cases}
    \pi_{\theta_T}(\bs_t) +\epsilon_t   & \mbox{with probability} \; \xi, \;\; \epsilon_t \sim \mathcal{N}(0, \sigma_k^2 \cdot I) \\
    \pi_{\theta_T}(\bs_t)               & \mbox{otherwise (noise-free)}
    \end{cases}
\end{equation}
with noise $\epsilon_t$ sampled from a multivariate Gaussian with covariance matrix $\sigma_k^2 \cdot I$.
The purpose of sampling $\sigma_k$ from a uniform distribution with range $(0, \Lambda)$ is to ensure a sufficient spread of episodes with very little noise from the teacher's final trained policy as well as those with heavier noise. Furthermore, the $\xi$ probability controls how many data tuples $(\bs, \ba, r, \bs')$ will have noise-free actions from the teacher. We experimented with different combinations of $\Lambda \in \{0.00, 0.25, 0.50, 0.75, 1.00, 1.25, 1.50\}$ and $\xi \in \{0.50, 0.75, 0.90\}$.

For the second way of generating teacher data, with uniform noise, we use one parameter, $\Xi$, which we use to sample $\xi_k \sim {\rm Unif}(0, \Xi)$ per episode $k$. Given this fixed $\xi_k$ throughout an episode, the teacher's rollout policy is defined as:
\begin{equation}
    \ba_t = 
    \begin{cases}
    \epsilon_t              & \mbox{with probability} \; \xi_k, \;\; \epsilon_t \sim {\rm Unif}(-1, 1) \\
    \pi_{\theta_T}(\bs_t)   & \mbox{otherwise (noise-free)}
    \end{cases}
\end{equation}
where with probability $\xi_k$, the teacher policy is ignored and we instead take steps in the environment emulator by using a purely uniform action, with each scalar component in $\ba_t$ sampled uniformly and independently from the full action ranges. The purpose of this is to inject stronger noise levels into the environment, and to potentially convey a wider range of states from which the (noise-free) teacher policy must recover.  We have experimented with $\Xi \in \{0.5, 1.0\}$.

From running TD3 offline on these datasets, we were generally unable to get good performance on any environment with the potential exception of HalfCheetah-v3, but even then, the policies only achieved roughly half the performance as compared to the teacher. We also attempted arranging the rollout data in a curriculum based on noise levels or reward, as well as including an analogous ``noise predictor'' as with the time predictor from Section~\ref{ssec:time-reward}, but none of these led to fruitful student learning results.
Inspecting the student training showed that TD3's Q-value estimates frequently diverged during training. These findings are also consistent with results from~\cite{BCQ_2019} and~\cite{behavior_regularized_2019}, which suggest that learning offline from data generated via policy rollouts with noise is difficult without a specialized offline RL algorithm, and they motivate learning from logged environment interaction instead, where TD3 (and SAC) can have greater success learning offline as shown in this work.

\end{document}